\documentclass[abstracton]{scrartcl}
\usepackage{ifluatex}
\ifluatex
	\usepackage[utf8]{luainputenc}
	\usepackage{luatexja-fontspec}
	\setmainjfont{FandolSong}
\else
	\usepackage[utf8]{inputenc} 
\fi

\usepackage[unicode=true,
bookmarks=true,bookmarksnumbered=false,bookmarksopen=false,
breaklinks=true,pdfborder={0 0 0}]
{hyperref}
\hypersetup{pdftitle={Deterministic Pod Repositioning Problem in Robotic
	Mobile Fulfillment Systems},
	pdfauthor={Ruslan Krenzler, Lin Xie, Hanyi Li},
	pdfsubject={logistics},
	pdfkeywords={logistics, applied combinatorial optimization, robotic mobile fulfillment system, warehouse},
	colorlinks=true, 
	linkcolor=blue,
	urlcolor=blue,
	citecolor=blue,
	draft=false
}
\usepackage{color}
\usepackage{natbib}
\usepackage{float}
\usepackage{subfig}
\usepackage{mathtools}
\usepackage{amsmath}
\usepackage{amsthm}
\usepackage{amssymb}
\usepackage{graphicx}
\usepackage{algorithm}
\usepackage{algpseudocode}
\usepackage{enumitem}
\usepackage{amsmath}
\usepackage{appendix} 
\usepackage{varioref} 
\usepackage{refstyle}
\usepackage{authblk} 

\floatstyle{ruled}
\newfloat{algorithm}{tbp}{loa}
\providecommand{\algorithmname}{Algorithm}
\floatname{algorithm}{\protect\algorithmname}

\theoremstyle{definition}
\ifx\thechapter\undefined
\newtheorem{example}{\protect\examplename}
\else
\newtheorem{example}{\protect\examplename}[chapter]
\fi
\theoremstyle{remark}
\ifx\thechapter\undefined
\newtheorem{rem}{\protect\remarkname}
\else
\newtheorem{rem}{\protect\remarkname}[chapter]
\fi

\theoremstyle{remark}
\ifx\thechapter\undefined
\newtheorem{techrem}{Technical remark}
\else

\fi

\theoremstyle{plain}
\ifx\thechapter\undefined
\newtheorem{prop}{\protect\propositionname}
\else

\fi
\theoremstyle{plain}
\ifx\thechapter\undefined
\newtheorem{assumption}{\protect\assumptionname}
\else

\fi
\usepackage{subfig}
\makeatother

\providecommand{\examplename}{Example}
\providecommand{\propositionname}{Proposition}
\providecommand{\remarkname}{Remark}

\global\long\def\sysstateset{S_{\text{sys}}}
\global\long\def\actionset{A}
\global\long\def\placeset{P}
\global\long\def\podset{S_{\text{pod}}}
\global\long\def\stationset{S_{\text{stn}}}
\global\long\def\stationmaxsize{M_{\text{stn}}}
\global\long\def\queue#1{[#1]}
\global\long\def\zerobasedlist#1{[#1]}
\global\long\def\storagestate{s_{\text{strg}}}
\global\long\def\storagestatepr{s_{\text{strg}}^{\prime}}
\global\long\def\storagestatet#1#2{s_{\text{strg},#1}^{#2}}
\global\long\def\enque{\text{enq}}
\global\long\def\deque{\text{deq}}
\global\long\def\nextpod{b}
\global\long\def\nextpodt#1#2{\nextpod_{#1}^{#2}}
\global\long\def\nextpods{\mathbf{b}}
\global\long\def\nextpodt#1#2{\nextpod_{#1}^{#2}}
\global\long\def\action{a}
\global\long\def\actiont#1#2{\action_{#1}^{#2}}
\global\long\def\coststostation{c_{\text{to stn}}}

\global\long\def\costsfromstation{c_{\text{from stn}}}
\global\long\def\maxtime{N}
\global\long\def\stationstate#1{q_{\text{stn}}(#1)}
\global\long\def\stationstatepr#1{q_{\text{stn}}^{\prime}(#1)}
\global\long\def\stationstatet#1#2#3{q_{\text{stn},#1}^{#2}(#3)}
\global\long\def\sysstate{s_{\text{sys}}}
\global\long\def\sysstatepr{s_{\text{sys}}^{\prime}}
\global\long\def\sysstatet#1#2{s_{\text{sys},#1}^{#2}}
\global\long\def\timespace{T_{\text{time}}}
\global\long\def\dponestagecosts{c}
\global\long\def\dpNstagecosts{C}
\global\long\def\stationweight#1{w_{\text{stn},#1}}
\global\long\def\StationRelativeFrequency#1{v_{\text{stn},#1}}
\global\long\def\podweight#1{w_{\text{\ensuremath{\pod}},#1}}

\global\long\def\podprojection#1{#1|_{\text{pod}}}
\global\long\def\stationprojection#1{#1|_{\text{station}}}
\global\long\def\podelement{h}
\global\long\def\stationelement{s}
\global\long\def\placeelement{p}
\global\long\def\AverageCosts{c_{\text{avg}}}

\global\long\def\DeterministicCosts{c_{\text{decis}}}
\global\long\def\FixPlaces{\action_{\text{fix}}}

\global\long\def\Indicator#1{1_{\{#1\}}}
\global\long\def\FromStationCount{f_{\text{from stn}}}
\global\long\def\ToStationCount{f_{\text{to stn}}}
\global\long\def\NumTimeIntervals{K}
\global\long\def\SetOfTimeIntervals{\mathcal{I}}

\global\long\def\FromStation#1{s_{\text{from}}(#1)}
\global\long\def\ToStation#1{s_{\text{to}}(#1)}

\global\long\def\indicator#1{1_{\{#1\}}}
\global\long\def\InitialBusyEnd#1{B_{\text{init},#1}}
\global\long\def\BusyEnd#1{B_{\text{end},#1}}
\global\long\def\BusyStart#1{B_{\text{start},#1}}
\newcommand{\ProblemName}{Pod Repositioning}
\renewcommand{\pod}{pod}
\newcommand{\pods}{pods}

\newcommand{\Pods}{Pods}

\newcommand{\outputStation}{pick station}
\newcommand{\outputStationAdj}{pick-station} 
\newcommand{\outputStations}{pick stations}

\newcommand{\systemFiveHundreds}{medium-size test system}
\newcommand{\SystemFiveHundreds}{Medium-size test system}

\algnewcommand{\LonelyComment}[1]{\Statex \(\triangleright\) #1}
\algnewcommand\algorithmicforeach{\textbf{for each}}
\algdef{S}[FOR]{ForEach}[1]{\algorithmicforeach\ #1\ \algorithmicdo}

\newcommand{\Rawsimo}{RAWSim-O}

\begin{document}

\title{Deterministic Pod Repositioning Problem in Robotic Mobile Fulfillment Systems}
\date{October 9, 2018}

\renewcommand\Affilfont{\small \itshape}

\author[a]{Ruslan Krenzler}
\affil[a]{Leuphana University of L\"uneburg, Universit\"atsallee 1, 21335 L\"uneburg }
\author[a]{Lin Xie}
\author[b]{Hanyi Li}
\affil[b]{Hanning ZN Tech Co., Ltd.}
\renewcommand\Authands{ and }

\maketitle
\begin{abstract}
In a robotic mobile fulfillment system, robots bring shelves, called
\textit{pods}, with storage items from the storage area to pick stations.
At every pick station there is a person -- the picker -- who takes
parts from the pod and packs them into boxes according to orders.
Usually there are multiple shelves at the pick station. In this case,
they build a queue with the picker at its head. When the picker does
not need the pod any more, a robot transports the pod back to the
storage area. At that time, we need to answer a question: ``Where
is the optimal place in the inventory to put this pod back?''. It
is a tough question, because there are many uncertainties to consider
before answering it. Moreover, each decision made to answer the question
influences the subsequent ones. The goal of this paper is to answer
the question properly. We call this problem the {\ProblemName} Problem
and formulate a deterministic model. This model is tested with different
algorithms, including binary integer programming, cheapest place,
fixed place, random place, genetic algorithms, and a novel algorithm
called \emph{tetris}.


\begingroup
\renewcommand\thefootnote{}%
\footnote{ORCID IDs and email addresses:\\
	Ruslan Krenzler \includegraphics[width=0.8em]{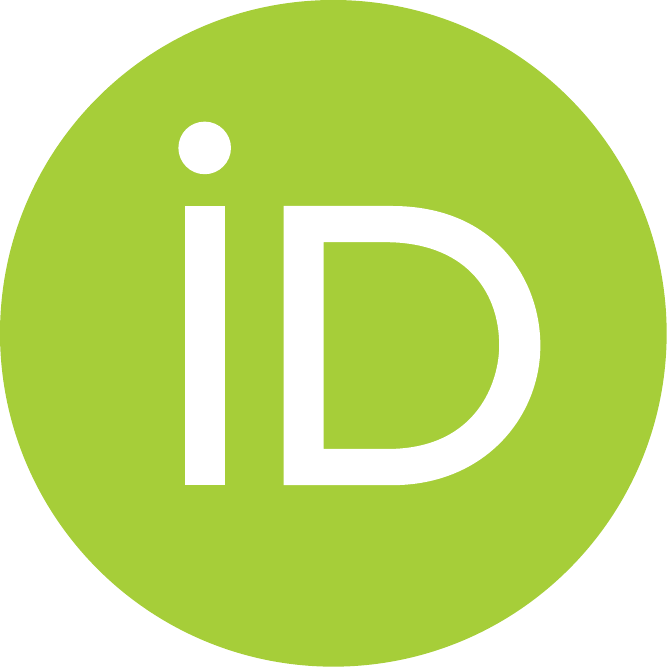}\hspace{0.4em}\url{https://orcid.org/0000-0002-6637-1168},
	ruslan.krenzler@leuphana.de,\\
	Lin Xie \includegraphics[width=0.8em]{orcid_icon}\hspace{0.4em}\url{https://orcid.org/0000-0002-3168-4922},
	xie@leuphana.de}%

\addtocounter{footnote}{-1}%
\endgroup

\end{abstract}

\newpage

\tableofcontents{}
\newpage

\section{Introduction}\label{sec:intro}
Robotic mobile fulfillment systems (RMFS) are a new type of warehousing
system which is becoming popular due to increasing growth in the e-commerce
sector: see \citet*{banker:2016}. In such systems, robots carry mobile
shelves -- called \emph{{\pods} --} with items from the storage
area to human operators -- the pickers -- at {\outputStations}.
At each station, the picker picks the items according to pick orders.
A pick order is basically the content of your shopping cart, when
you buy something online. Usually, there are multiple {\pods} at
every {\outputStation}. In this case, they build a queue with the
picker at its head. After the picker has picked all the items, the
robot carries the {\pod} back to the storage area and selects another
{\pod}.

Such a fulfillment system contains many different operational decision
problems, see \citet*[Section 2.1]{Merschformann-xie-li:2017}. In
this work, we concentrate on one question: ``Where to put the pod
back to, after the {\outputStation}?'' To support an intuitive understanding
we sketch this problem with the help of the following small example.

\paragraph*{Example}

\begin{figure}
\begin{centering}
\includegraphics{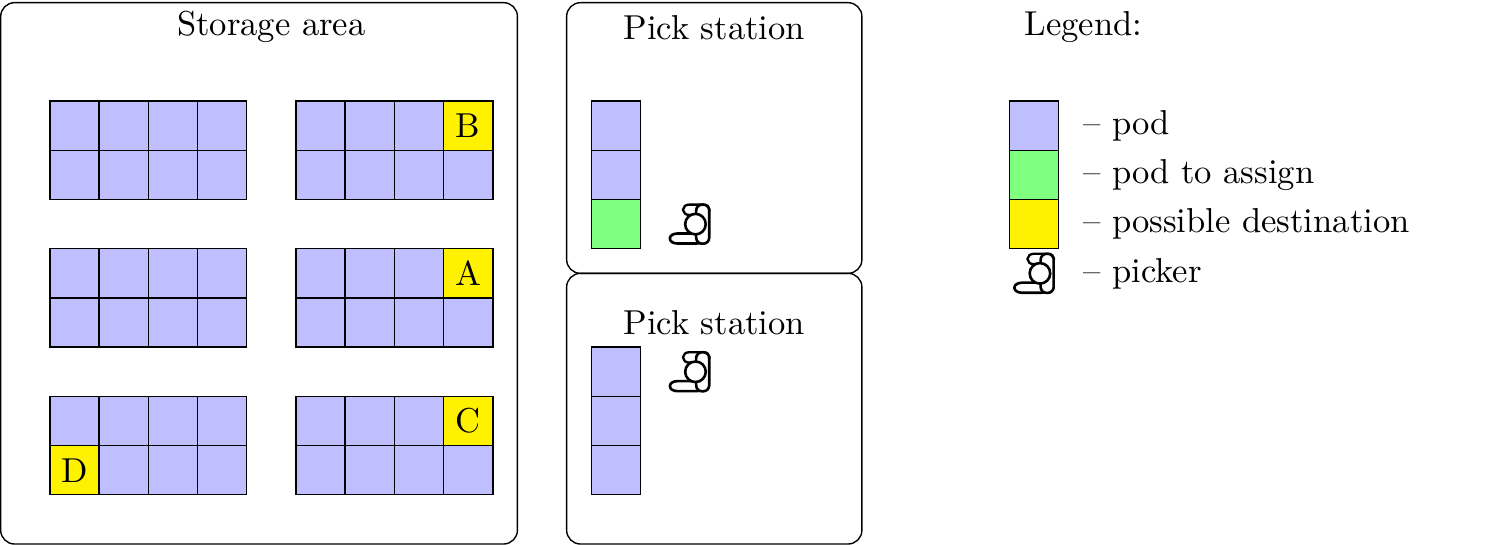}
\par\end{centering}
\caption{\label{fig:decision-problem}Possible places  to move the green {\pod}
to.}
\end{figure}

\Figref{decision-problem} shows possible destinations -- A, B, C
and D -- for a {\pod} (marked green) which leaves the upper {\outputStation}.
We want to minimize the traveling distance of the robots. Depending
on the situation, each of the places A, B, C and D can be optimal:
\begin{itemize}
\item A is optimal if we want to move the {\pod} to the closest place.
\item B is optimal if we want to move the {\pod} later to the upper {\outputStation}.
\item C is optimal if we want to move the {\pod} later to the lower {\outputStation}.
\item D is optimal if we will not use the green {\pod} any-more. Then we
can use places A, B, and C for other more frequently used {\pods}.
\end{itemize}

We call this decision problem \emph{the {\ProblemName} Problem} (in
short: PRP). There is only one publication about this problem, since
this problem appears firstly with an RMFS: In \citet*{Merschformann:2018},
Merschformann analyzed PRP in regard to passive and active repositioning.
Passive repositioning means that the pods find a better location after
visiting a {\outputStation}; while active repositioning means that
the pods move from their current locations to better ones without
visiting any {\outputStation} in between. That paper concentrates
on active repositioning. The author uses simulations to investigate
the effects of three different repositioning mechanisms. He shows
that the best strategy is to use the nearest available storage location
in terms of the path time. He also shows that the positive effects
can be achieved during a nightly down period, but not in all cases,
especially if the passive repositioning already keeps the storage
area sorted well. Moreover, the active repositioning during the nightly
down period faces additional costs such as energy costs. The PRP is
a problem of operations planning. An overview of decision problems
in an RMFS is listed in \citet*{azadeh-de-koster-roy:2018}, where
publications about other problems are also mentioned, such as system
analysis and design optimizations.

The contribution of our work are listed as follows:
\begin{itemize}
\item Our work is the first paper to concentrate on the passive case of
PRP, because we expect better performance of an RMFS and save additional
repositioning costs.
\item We create a deterministic mathematical model to better understand
PRP\@.
\item We analyze classical solving methods: binary integer programming,
cheapest place, fixed place, random place, iterative binary integer
programming, and genetic algorithms.
\item We create a heuristic to solve large instances quickly.
\item From the model, we derive some general rules, which one can apply
in the real world.
\item We critically discuss if and how our algorithms can be implemented
in the real world. 
\end{itemize}
The remainder of the paper is structured as follows. First, we define
the process of an RMFS and the decision problems in it in \Secref{rmfs}.
After that, we formulate the PRP as a deterministic model under some
simplifications in \Secref{game}. We introduce two test systems for
algorithm analysis in \Secref{test-systems} and provide suitable
exact and heuristic solution approaches in \Secref{algorithms}. We
briefly analyze the computational results in \Secref{results} and
discuss real-world implementation of the solvers in \Secref{problem-with-multiple-stations}.
Finally, \Secref{conclusion} concludes the paper.

\section{The robotic mobile fulfillment system}\label{sec:rmfs}
\begin{figure}[h]
\begin{centering}
\includegraphics[width=1\textwidth]{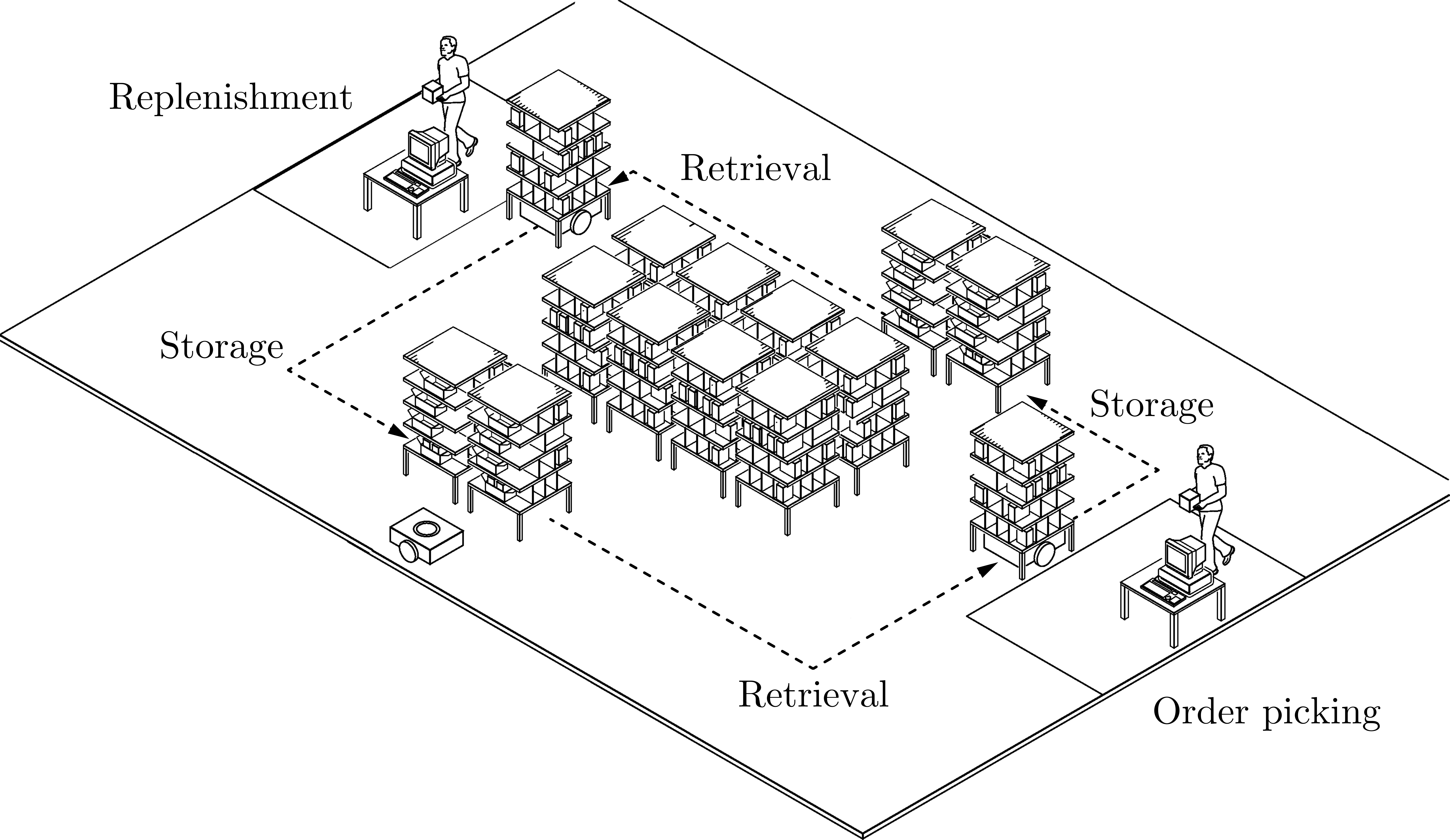}
\par\end{centering}
\centering{}\caption{The central process of an RMFS (see \citet*{Hoffman:2013}).\label{fig:storage_retrieval_process} }
\end{figure}

Before we explain the essential processes in an RMFS, we define several
terms which we did not mention in the introduction:
\begin{itemize}
\item \emph{workstations} -- places where the persons interact with the
pods; a special type of a workstation is a {\outputStation} 
\item \emph{replenishment station} -- a workstation where the persons replenish
the pods
\item stock keeping unit (\textit{SKU})
\item an \textit{order line} including one SKU with number
\item a \textit{pick order} including a set of order lines from an order
of a customer
\item a \textit{replenishment order} consisting of a number of physical
units of one SKU 
\end{itemize}
The processes of an RMFS are illustrated in \Figref{storage_retrieval_process}.
The robots carry {\pods} between the storage area and workstations.
Two processes are included: 
\begin{itemize}
\item \textit{retrieval process}: After a replenishment order has arrived,
robots carry selected {\pods} to a replenishment station, where the
units are stored in these pods. We assume that a shared storage policy
is applied (such as in \citet*{bartholdi-hackman:2016}), which means
SKUs of the same type are not stored together in a unique pod, but
are spread over several pods. 
\item \textit{storage process}: After a pick order has arrived, we calculate
which pods we need to process the order lines. Robots carry required
pods to a {\outputStation}, where the picker picks the SKUs according
to the order lines. We assume it is unlikely that a pick order can
be completed with only one pod, unless there is only one order line
or the association policy was applied in the retrieval process (in
other words, all SKUs are stored together in one pod, if they are
often ordered together by the same customer). 
\end{itemize}
Then, after a pod has been processed at one or more stations, it is
brought back to the storage area. 

In the following sections we will use following notations:
\begin{itemize}
\item $\mathbb{N}$; natural numbers $1,2,3,\ldots$; $\mathbb{N}_{0}:=\mathbb{N}\cup\{0\}$;
$\mathbb{R}^{+}$ positive real numbers
\item $[a,b)$; half opened interval $\{x\in\mathbb{R}:a\leq x<b\}$
\item $a_{t}$ some value $a$ at time $t$, $a_{t}^{N}$ sequence of values
$(a_{t},a_{t+1},a_{t+2},\ldots a_{N})$
\item $\subset$ subset; it also includes the equal set
\item $\Indicator{expression}$; is an indicator function; it is $1$ if
the \emph{expression} is true and $0$ otherwise
\item $[a_{0},a_{1},\ldots]$; ordered set (list). The element in the beginning
has index \emph{zero}.
\end{itemize}

\section{Warehouse game}\label{sec:game}
A real robotic warehouse is very complex. Even its simulation is still
too complex to be completely modeled mathematically. Therefore we
create a simplified mathematical model which reflects only the parts
of an RMFS essential for the {\ProblemName} Problem.

\subsection{Simplifications}

\paragraph{No explicit pick orders}

In the real world, a customer orders one or multiple items from an
online shop. In the shop's warehouse, these items are in a single
{\pod} or in multiple {\pods} in the storage area. The warehouse
commands the robots to move the corresponding {\pods} to a {\outputStation}
where a picker will pack the items for the customer. For the {\ProblemName}
Problem it is not important what items the customer has ordered and
what items are in the {\pods}. For us, only the direct consequence
of the order is important: at some point in time a particular {\pod}
is at the {\outputStation} and then it returns back to the storage
space.

\paragraph{No explicit robots}

In the real world, we need to know exactly where each particular robot
is and which {\pod} it carries. For our mathematical model, the only
important properties are the position of the robots' pods in the {\outputStationAdj}
queue and the number of robots. This number is equal to the maximal
number of {\pods} at all {\outputStations}. 

\paragraph{No replenishment}

In this model we neglect replenishment of the {\pods}, because it
happens much less frequently than emptying of the {\pods} at the
{\outputStations}.

\paragraph{Keep queues full\label{page:keep-queues-full}}

We require our algorithms to keep the queues at the {\outputStations}
as full as possible. The reason for this requirement is: When we have
many {\pods} at the {\outputStations}, we have many free places
in the storage area. More free places in the storage area mean more
possibilities to select cheaper places and to reduce the costs.

\paragraph{Time discretization}

\begin{figure}
\begin{centering}
\subfloat[\label{fig:real-starting-state}Starting state.]{\centering{}\includegraphics{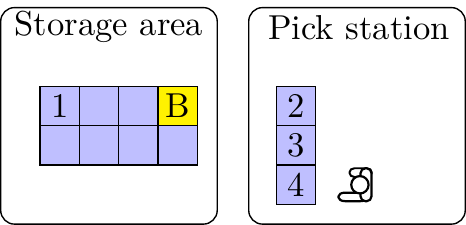}}\qquad{}\subfloat[\label{fig:real-intermediate-state}Intermediate state.]{\centering{}\includegraphics{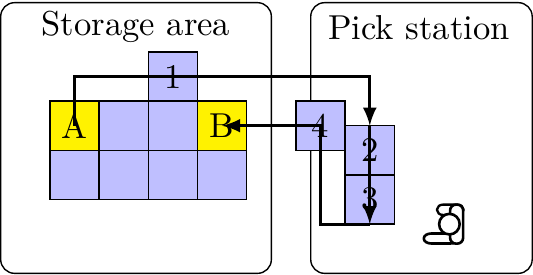}}
\par\end{centering}
\begin{centering}
\subfloat[\label{fig:real-final-state}Final state.]{\centering{}\includegraphics{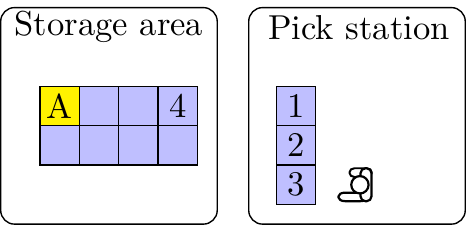}}
\par\end{centering}
\centering{}\caption{\label{fig:time-discretization}Time discretization. We start the
system in a state shown in \Figref{real-starting-state}. In continuous
time, {\pod} 1 is leaving the storage area. At the same time, {\pod}
4 is leaving the {\outputStation} and the remaining {\pods}, 2 and
3, are moving within the {\outputStationAdj} queue towards the picker.
After some time, the system arrives at the state shown in \Figref{real-final-state}.
In our discrete-time model, we go directly from the state in \Figref{real-starting-state}
into the state in \Figref{real-final-state}.}
\end{figure}

In the real world, many {\pods} are moving simultaneously between
the storage area and the {\outputStations} and within the {\outputStations}.
To simplify the model, we split the time into discrete steps and we
combine different movements of the {\pod} into a single step. Each
time step combines the movement of a {\pod} from the storage area
to the {\outputStationAdj} queue, movement within the {\outputStationAdj}
queue, and the movement of a {\pod} out of the station. See \Figref{time-discretization}.

\begin{rem}
\label{rem:have-free-places}In the ``Keep queues full'' requirement,
we have showed that free places in the storage area are important
for optimization. In a real warehouse, there are at least two approaches
how to create them: The first one is to have more places than the
total number of {\pods} in the system. The disadvantage of this approach
is long traveling distances of the {\pods} and costs for the unused
space in the storage. The second approach is to keep many {\pods}
out of the storage area by moving them into long queues at the picking
station or into a \emph{drift space. For drift space,} see Amazon's
patent \citet*{patent-wurman-brunner-barbehenn:2013}. The disadvantage
of this approach is the requirement to have large number of robots,
that stays all the time under the {\pods}.
\end{rem}

\subsection{Parameters and states}
\begin{itemize}
\item Time space $\timespace=\{0,1,2,\ldots,\maxtime\}\subset\mathbb{N}_{0}$,
where $\maxtime$ is the maximal number of steps.
\item Finite set of {\pods} $\podset=\{1,2,\ldots\}\subset\mathbb{N}$. 
\item Finite set of places $\placeset=\{1,2,\ldots\}\subset\mathbb{N}$.
\item Finite set of stations $\stationset=\{1,2,\ldots\}\subset\mathbb{N}$.
Each station $s\in\stationset$ contains maximal $\stationmaxsize(s)$
{\pods}.
\item $\stationstate{\stationelement}=$ $\queue{\podelement_{0},\podelement_{1},\podelement_{2},\ldots}$
of the station $s$. It is a first-in, first-out (FIFO) queue with
{\pod} $h_{0}\in\podset$ at the head, followed by {\pod} $h_{1}\in\podset$,
and so on. 
\begin{itemize}
\item $\stationstate{\stationelement,i}$ is the $i$-th element of the
queue counted from the head. The head has index $0$.
\item $\enque$ is the enqueue operator: 
\[
\hspace{-1cm}\enque\left(\queue{\podelement_{0},\podelement_{1},\ldots,\podelement_{L}},\podelement_{L+1}\right)=\begin{cases}
\queue{\podelement_{0},\ldots,\podelement_{L},\podelement_{L+1}}, & \text{if queue was not full}\\
\queue{\podelement_{1},\ldots,\podelement_{L},\podelement_{L+1}}, & \text{if queue was full}
\end{cases}
\]
\item $\deque$ is the dequeue operator:
\[
\hspace{-1cm}\deque\left(\queue{\podelement_{0},\podelement_{1},\ldots,\podelement_{L}}\right)=\queue{\podelement_{1},\ldots,\podelement_{L}}.
\]
\end{itemize}
\item $\storagestate:\placeset\rightarrow\podset\cup\{0\}$, $\storagestate(\placeelement)=\podelement\in\podset$
means place $\placeelement$ is occupied by {\pod} $\podelement$.
$\storagestate(\placeelement)=0$ means place $\placeelement$ in
the storage area is free. The storage can be represented as an indexed
family with index set $\podset$. For example $(4,5,0,0,1,0)$ means
there is {\pod} $4$ at place $1$, {\pod} $5$ at place 2, and
{\pod} $1$ at place $5$; all other places are free.
\item $\nextpodt t{}$ describes {\pod} departure from the storage space.
A value $\nextpodt t{}=(\podelement,\stationelement)$ means that
at time $t+1$ a {\pod} $\podelement$ will go to station $\stationelement$.
\begin{itemize}
\item $\podprojection{\nextpodt t{}}$ is the {\pod} component of $\nextpodt t{}$.
This means it is $\podelement$ for $\nextpodt t{}=(\podelement,\stationelement)$.
\item $\stationprojection{\nextpodt t{}}$ is the station component of $\nextpodt t{}$.
This means it is $\stationelement$ for $\nextpodt t{}=(\podelement,\stationelement)$.
\item $\nextpods$ is a sequence of all future departures. At time $t$
it is $\nextpodt t{N-1}:=(\nextpodt t{},\nextpodt{t+1}{},\ldots,\nextpodt{N-1}{})$.
\end{itemize}
\end{itemize}
\begin{itemize}
\item The state of the system $\sysstate$ consists of the state of the
storage $\storagestate$, current state of the station queues $\left(\stationstate{\stationelement}\right)_{\stationelement\in\stationset}$
and the future departures $\nextpods$.
\item $\action\in\placeset\cup\{0\}$ is an action in the sense of \emph{discrete-time
dynamic programming}. It is a decision to move a {\pod} from the
head of the queue at a station $\stationprojection{\nextpod}$ to
some free place in the storage. $a=0$ means no {\pod} leaves the
stations.
\item $\actionset$ is the action space. $\actionset:=\placeset\cup\{0\}$.
\item $D(\sysstate)\subset\actionset$ is a set of admissible actions for
system state $\sysstate$.
\end{itemize}

\subsection{Dynamics\label{subsec:Dynamics}}

We describe the changes of the system over the time with discrete-time
dynamic programming. See \citet*[Chapters 2 and 3]{hinderer-rieder-stieglitz-2016}. 

Following the requirement of full {\outputStationAdj} queues \vpageref{page:keep-queues-full},
during the first time steps we fill the queues. Formally, assume at
time $t$ the system is in state $\sysstate=\left(\storagestate,\left(\stationstate{\stationelement}\right)_{\stationelement\in\stationset},\nextpodt t{N-1}\right)$.
We select a {\pod} $\podprojection{\nextpodt t{}}$ at place $\placeelement$
and put it into not-full station $\stationelement:=\stationprojection{\nextpodt t{}}$.
No {\pod} is allowed to leave the queue and the action $a_{t}$ must
be zero. At time $t+1$, the station changes to the state $\stationstatepr{\stationelement}:=\enque\left(\stationstate{\cdot},\podprojection{\nextpodt t{}}\right)$
and the storage area gets a gap at position $\placeelement$ and changes
to $\storagestatepr:=(\ldots,\underbrace{0}_{\mathclap{\text{at position }\placeelement}},\ldots)$.

Most of the time the system runs with full queues. A newly arrived
{\pod} ``pushes'' another {\pod} out of this queue. More precisely,
assume at time $t$ a system is in state $\sysstate=\left(\storagestate,\left(\stationstate{\stationelement}\right)_{\stationelement\in\stationset},\nextpodt t{N-1}\right)$.
We select a {\pod} $\podprojection{\nextpodt t{}}$ and put it into
the corresponding station $\stationelement:=\stationprojection{\nextpodt t{}}$.
Because station $\stationelement$ is full, the {\pod} at the head
of queue $\stationstate{\stationelement}$ departs immediately and
station $s$ changes its state to $\stationstatepr{\stationelement}:=\enque\left(\stationstate{\stationelement},\podprojection{\nextpodt t{}}\right)$.
The remaining stations stay unchanged.

In the same instant of time $t$, an action $a$ decides where to
put {\pod} $\podelement:=\stationstate{\stationelement,0}$ from
the head of station $\stationelement$. This action can choose only
free places in storage $\storagestate$ and the place that has been
newly freed by the departed {\pod} $\podprojection{\nextpodt t{}}$:
\[
a\in D(\sysstate):=\{\placeelement\in\placeset:\underbrace{\storagestate(\placeelement)=0}_{\mathclap{\text{free places}}}\vee\underbrace{\storagestate(\placeelement)=\podprojection{\nextpodt t{}}}_{\mathclap{\text{newly freed place}}}\}.
\]
The storage state changes from $\storagestate=(\ldots,\underbrace{0}_{\mathclap{a\text{-th position}}},\ldots)$
or $\storagestate=(\ldots,\underbrace{\podprojection{\nextpodt t{}}}_{\mathclap{a\text{-th position}}},\ldots)$
to $\storagestatepr:=(\ldots,\underbrace{\podelement}_{\mathclap{a\text{-th position}}},\ldots)$.
For the following system state, we do not need information about {\pods}
departed previously to time $t$, and the sequence of the future departures
changes to $b_{t+1}^{N-1}$. The new system state is $\sysstatepr:=\left(\storagestatepr,\left(\stationstatepr{\stationelement}\right)_{\stationelement\in\stationset},\nextpodt{t+1}{N-1}\right)$.

We summarize the dynamics described in this section as a \emph{transition
function}
\begin{equation}
\begin{array}{cc}
T:\sysstateset\times\actionset & \longrightarrow\sysstateset,\\
(\sysstate,a) & \mapsto\sysstatepr.
\end{array}\label{eq:deterministic-transition-function}
\end{equation}

\begin{example}
Given six {\pods} $\podset=\{1,2,3,4,5,6\}$, two stations $S=\{1,2\}$.
Each station has maximal length $\stationmaxsize(1)=\stationmaxsize(2)=2$.
The system starts at time $t=0$ and ends at time $t=N:=2$. The initial
system state is 
\[
\sysstatet 0{}:=(\storagestatet 0{},\stationstatet 0{}1,\stationstatet 0{}2,b_{0}^{1})
\]
with $\storagestatet 0{}:=(1,0,3,0,0,0)$, $\stationstatet 0{}1=\queue{5,2}$,
$\stationstatet 0{}2=\queue{4,6}$ and $b_{0}^{1}=\left((3,2),(1,2)\right)$.
The action at time $0$ is $\actiont 0{}=3$.

Because of departure $b_{0}=(3,2)$, the system puts {\pod} $3$
into station $2$. The action $\actiont 0{}=3$ moves {\pod} $4$
from the head of  queue $\stationstatet 0{}2=\queue{4,6}$ to place
$3$ in the storage state $\storagestatet 0{}$. The following state
$\sysstatet 1{}$ then becomes
\[
\sysstatet 1{}:=(\storagestatet 1{},\stationstatet 1{}1,\stationstatet 1{}2,b_{1}^{1})
\]
with $\storagestatet 1{}:=(1,\underbrace{4}_{\mathclap{\text{arrived}}},\underbrace{0}_{\mathrlap{3\text{ departed}}},0,0,0)$,
$\stationstatet 1{}1=\queue{5,2}$, $\stationstatet 1{}2=\queue{\underbrace{6}_{\mathclap{\text{moved up}}},3}$
and $b_{1}^{1}=\left((1,2)\right)$.
\end{example}

\subsection{Costs}

\begin{figure}[h]
\centering{}\includegraphics{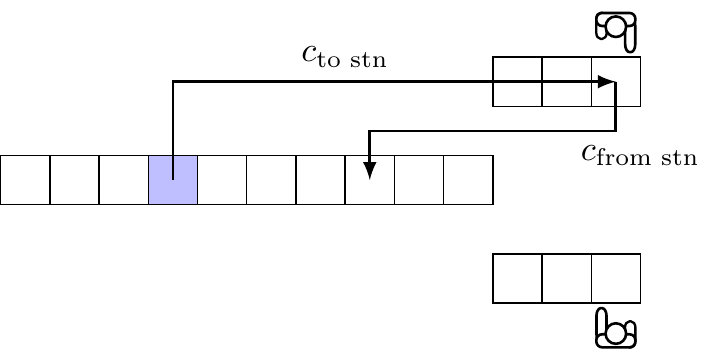}\caption{Two types of costs, caused by the {\pod} movements in a single time
step.\label{fig:costs}}
\end{figure}

Each movement of a {\pod} causes non-negative costs. They depend
on the location of the {\pod} and consist of two parts -- see \Figref{costs}:
\begin{enumerate}
\item Moving a {\pod} from the storage to a station costs
\begin{equation}
\coststostation:\placeset\times\stationset\longrightarrow\mathbb{R}_{0}^{+}.
\end{equation}
\item Moving a pod from a station to the storage costs
\begin{equation}
\costsfromstation:\stationset\times\placeset\longrightarrow\mathbb{R}_{0}^{+}.
\end{equation}
\end{enumerate}
Using the terminology of dynamical programming, the costs for an action
$a$ applied to a system state $\sysstate=\left(\storagestate,\left(\stationstate{\stationelement}\right)_{\stationelement\in\stationset},\nextpodt t{N-1}\right)$
are described by a function
\begin{align*}
\dponestagecosts & :\actionset\times\sysstateset\longrightarrow\mathbb{R}.
\end{align*}
If departure $b_{t}$ sends a {\pod} $\podelement$ from place $\placeelement$
to station $\stationelement$, it holds
\begin{equation}
\dponestagecosts(a,\sysstate)=\coststostation(\placeelement,\stationelement)+\costsfromstation(\stationelement,a).\label{eq:deterministic-dp-cost-function}
\end{equation}
 The total costs for a decision sequence $y:=a_{0}^{N-1}$ for the
initial state $\sysstatet 0{}$ correspond to the $N$-stage objective
function 
\begin{equation}
\dpNstagecosts_{Ny}(\sysstatet 0{}):=\sum_{t=0}^{N-1}\dponestagecosts(a_{t},\sysstatet t{})+C_{0}(\sysstatet N{}),\label{eq:N-stage-objective-function}
\end{equation}
where $\sysstatet t{}$ are generated from $\sysstatet 0{}$ by $y$
using the transition function $T$ from (\ref{eq:deterministic-transition-function})
and $C_{0}$ is the \emph{terminal cost function}.

In our model we work with two types of terminal costs:
\begin{enumerate}
\item Zero terminal costs. We use this for most algorithms when $N$ is
large and therefore $C_{0}$ is negligible.
\item (Estimated) future costs, when $N$ is small. It forces us to make
good strategic decisions before we stop the system and resume it from
the last state. We use these costs in the iterative Binary Integer
Programming (BIP): see \Remref{iterative-bip}.
\end{enumerate}
Other terminal costs are also possible. For example:
\begin{enumerate}[resume]
\item The minimal costs of carrying all the {\pods} in the output station
back to the storage place.
\end{enumerate}
\begin{rem}
The more general form of an $N$-stage objective function is\\
$\dpNstagecosts_{Ny}(\sysstatet 0{}):=\sum_{t=0}^{N-1}\beta^{t}\dponestagecosts(a_{t},\sysstatet t{})+\beta^{N}C_{0}(\sysstatet N{})$
with \emph{discount factor} $\beta\in\mathbb{R}^{+}$. For our analysis,
the simpler version (\ref{eq:N-stage-objective-function}) is sufficient.
\end{rem}

\subsection{Binary integer programming\label{subsec:binary-integer-programming}}

In the binary formulation of the problem, we mainly focus on the case
when the terminal costs are zero. That is $C_{0}\equiv0$.

\paragraph*{Decision variables}

$x_{t\placeelement}\in\{0,1\}$. $x_{t\placeelement}=1$ means at
time $t\in\timespace\setminus\{N\}$ we decide to put the current
{\pod} in place $\placeelement\in P$. In other words $x_{tp}$ corresponds
to action $a_{t}$:
\[
x_{tp}=1\Longleftrightarrow a_{t}=p.
\]

\paragraph*{Parameters}

At time $t\in\timespace\setminus\{N\}$ we decide to put a {\pod}
into the storage area. This {\pod} leaves station $\FromStation t$.
It arrives into the storage area at time $\BusyStart t:=t+1$ and
leaves it at time $\BusyEnd t\in\timespace$. Hence, the {\pod} stays
there for a time interval $[\BusyStart t,\BusyEnd t)$. We set $\BusyEnd t:=\max(\timespace)+1$
if the {\pod} does not leave within the timespan $\timespace$. Later,
 the {\pod} leaves the storage area and goes to station $\ToStation t$.
We set $\ToStation t$ equal to zero when the {\pod} does not leave
the storage area within the timespan $\timespace$.

$\InitialBusyEnd{\placeelement}$ is the first time a place $p$ becomes
free within the timespan $\timespace$.

\paragraph*{Cost function}

In the cost function, we ignore the costs of moving {\pods} to stations
at the beginning of the game, because we cannot influence them. These
are {\pods} whose position was not decided within the timespan $\timespace\setminus\{\maxtime\}$.

During the timespan $\timespace$ almost every {\pod} assigned to
a place will also leave this place during the same timespan $\timespace$.
The exceptions are the {\pods} which are marked by the parameter
$\ToStation t=0$. Thus, every decision $x_{tp}$ causes costs $\costsfromstation(\FromStation t,\placeelement)+\coststostation(\placeelement,\ToStation t)\cdot\indicator{\ToStation t\neq0}\cdot$
Consequently, the total costs for $C_{0}\equiv0$ are 

\begin{equation}
\sum_{t\in\timespace\setminus\{\maxtime\}}\sum_{\placeelement\in P}\left(\costsfromstation(\FromStation t,\placeelement)+\coststostation(\placeelement,\ToStation t)\cdot\indicator{\ToStation t\neq0}\right)\cdot x_{t\placeelement}.\label{eq:bin-objective}
\end{equation}

\paragraph*{Constraints}

A {\pod} must be assigned to only one place at time:

\[
\sum_{\placeelement\in\placeset}x_{t\placeelement}=1.
\]
The {\pod} may not arrive in a busy place:
\begin{equation}
\InitialBusyEnd{\placeelement}\leq\underbrace{\BusyStart t}_{=t+1}\cdot x_{t\placeelement}+M_{\text{big}}\cdot(1-x_{t\placeelement})\qquad\forall t\in\timespace\setminus\{\maxtime\},\placeelement\in\placeset,\label{eq:initial-conditions-busy-constraints}
\end{equation}
\begin{equation}
\BusyEnd{\tau}x_{\tau\placeelement}\leq\underbrace{\BusyStart t}_{=t+1}\cdot x_{t\placeelement}+M_{\text{big}}\cdot(1-x_{t\placeelement})\qquad\forall t\in\timespace\setminus\{N\},\tau<t,\placeelement\in\placeset,\label{eq:previous-decisions-busy-constraints}
\end{equation}
with a large value $M_{\text{big}}$. $M_{\text{big}}$ deactivates
both constrains $\eqref{eq:initial-conditions-busy-constraints}$
and (\ref{eq:previous-decisions-busy-constraints}) when at time $t$
we do not decide to put the current pod to place $\placeelement$
-- this is when $x_{t\placeelement}=0$. Constraints $\eqref{eq:initial-conditions-busy-constraints}$
consider the busy places resulting from the initial state of the system.
Constraints (\ref{eq:previous-decisions-busy-constraints}) consider
the busy places resulting from the previous decisions.

We can significantly reduce the number of constraints (\ref{eq:previous-decisions-busy-constraints})
if we consider only occupation time intervals which may overlap with
current decision $x_{t\placeelement}$. Formally, we need to consider
only $\tau$ with $\BusyEnd{\tau}>\BusyStart t$:
\begin{equation}
\begin{array}{cc}
\BusyEnd{\tau}x_{\tau\placeelement}\leq\underbrace{\BusyStart t}_{=t+1}\cdot x_{t\placeelement}+M_{\text{big}}(1-x_{t\placeelement})\qquad & \forall t\in\timespace\setminus\{N\},\tau<t,\\
 & \placeelement\in\placeset,\BusyEnd{\tau}>\BusyStart t.
\end{array}\label{eq:previous-decisions-busy-constraints-improved}
\end{equation}

\section{Test systems}\label{sec:test-systems}
To test all our algorithms, we will mainly use two systems: a small
system for qualitative analysis and a {\systemFiveHundreds} for more
realistic tests.

\subsection{Small test system\label{subsec:test-system-10}}

\begin{figure}[h]
\centering{}\subfloat[Layout, place positions and {\outputStationAdj} positions.]{\begin{centering}
\includegraphics{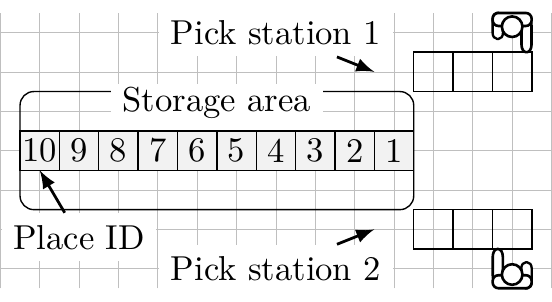}
\par\end{centering}
}\subfloat[Initial state and {\pods}.]{\centering{}\includegraphics{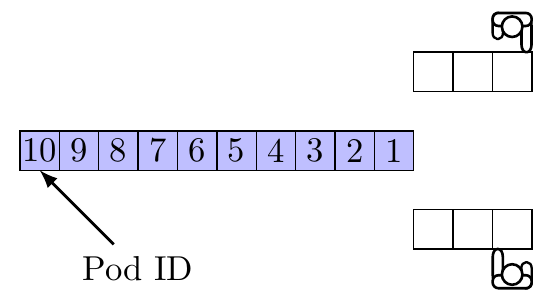}}\caption{Small system for qualitative analysis of algorithms.\label{fig:small-system-for-algorithm-analysis}}
\end{figure}

For the qualitative analysis of the Pod Storage Problem and its solving
algorithms we need a very small and simple system. See \Figref{small-system-for-algorithm-analysis}.
It has only 10 places and 10 \pods. The costs are Manhattan distances
between places in the storage area and places right in front of pickers.
For example $\coststostation(1,\stationelement)=5$ and $\costsfromstation(\stationelement,5)=9$
for every station $\stationelement\in\{1,2\}$. The storage space
is one-dimensional -- it makes it easy to identify the cheapest (nearest)
and the most expensive (most distant) places. We use two stations,
because according to our experience, a system with only one {\outputStation}
behaves very differently from a system with multiple {\outputStations}.
And, to our knowledge, there is no RMFS in the real world with only
one {\outputStation}. As a further simplification, we use equal {\outputStations}
which we place symmetrically to the storage area. Because of this
layout, the distance from any place in the storage area to every {\outputStation}
is the same.

To create a sequence of departures $\nextpods=\nextpodt 0{N-1}$,
we randomly choose a {\pod} and a {\outputStation}. In the real
world, some {\pods} are used more frequently than others. Also, due
to different pickers, some {\outputStations} have higher throughput
than others. To emulate this behavior we assign to each pod a positive
weight
\[
\podweight{\podelement}\in\mathbb{R}^{+},\qquad\podelement\in\podset,
\]
and to each station a positive weight
\begin{align*}
\stationweight{\stationelement} & \in\mathbb{R}^{+},\quad\stationelement\in\stationset, & \text{with } & \sum_{\stationelement\in\stationset}\stationweight{\stationelement}=1.
\end{align*}
To decide on the next departure $\nextpod$, we look at all {\pods}
in the storage $\storagestate$, and send {\pod} $\podelement$ to
station $\stationelement$ with probability $\stationweight{\stationelement}\frac{\podweight{\podelement}}{\sum_{k\in\storagestate}\podweight k}$.
Here, $\storagestate$ means the set of all {\pods} in the storage
area.

\begin{figure}
\centering{}\includegraphics{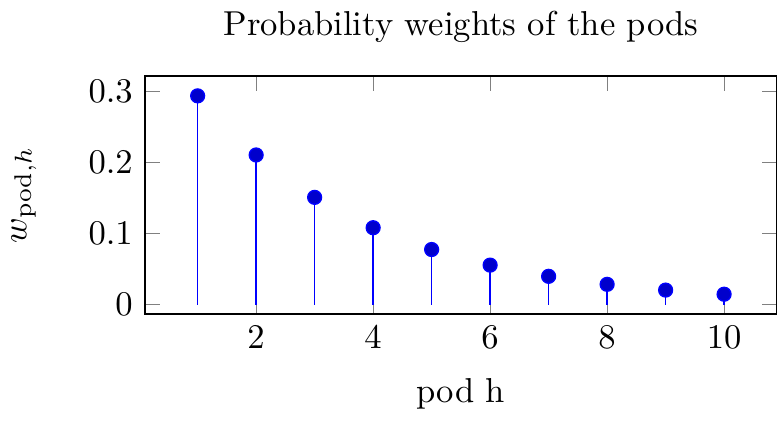}\caption{Probability weights of the {\pods} of a small system from \Figref{small-system-for-algorithm-analysis}.\label{fig:small-system-weights}}
\end{figure}

For our small system, we use {\pod} weights equal to the density
of the truncated geometric distribution
\[
\podweight{\podelement}=C^{-1}\podweight 1^{\podelement}\quad\text{with normalization constant }C:=\sum_{\podelement\in\podset}\podweight{\podelement}.
\]
We choose $\podweight 1$ in such a way that the weight of the most
frequently used {\pod} is 20 times\footnote{There is no particular reason why this number is 20 and not 19 or
$3\pi+\sqrt{2}$. The number should be high enough to have very different
frequencies of the {\pods}, but not too high, because we want to
use even the rarest {\pod} during the test run.} higher than the weight of the least frequently used {\pod}
\[
\frac{\podweight 1}{\podweight{10}}=20\Longrightarrow\podweight 1=0.29365446.
\]
See \Figref{small-system-weights}.

For our simple system we use equal station weights: $\stationweight 1=\stationweight 1=1/2$.
The number of time steps is $N=1000$ and we use the same sequence
$\nextpods$ for every algorithm.

We start the small system with pre-sorted {\pods}. A more frequently
used {\pod} is closer to the {\outputStations} than a less frequently
used {\pod}. 

\subsection{Storage-area chart}

\begin{figure}[h]
\begin{centering}
\includegraphics{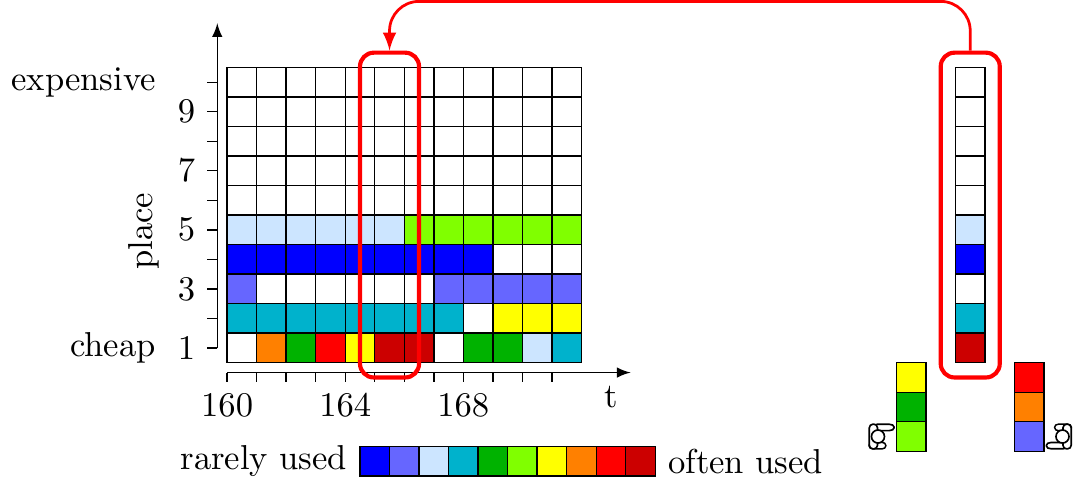}
\par\end{centering}
\caption{Section of storage-area chart. The red-framed section is storage state
at the time $165$ or during the time span $[160,173)$.\label{fig:storage-area-chart-explanation}}
\end{figure}

For visual analysis of algorithms in \Secref{algorithms} we will
use special graphs -- storage-area charts. They show how the small
system changes in time. The vertical axis shows places. See detailed
explanation in \Figref{storage-area-chart-explanation}. The horizontal
axis shows time. The colors show which {\pod} occupies the current
place. Every {\pod} has its own unique color all the time. We use
colors from dark blue to dark red: from least used to most used {\pods}.
If the place is empty, we mark it white.
\begin{rem}
The storage-area chart is a Gantt chart. The places are machines.
The not interrupted sequences of squares, which belongs to the same
{\pod}, are the jobs. The chart shows that {\ProblemName} Problem
is a special case of an NP-complete Problem\emph{ Interval Scheduling
with Required Jobs}, see \citet*[Subsection 2.2.1]{kolen-lenstra-papadimiriou-spieksma:2007}.
\end{rem}

\subsection{{\SystemFiveHundreds}\label{subsec:test-system-504}}

\begin{figure}
\centering{}\includegraphics{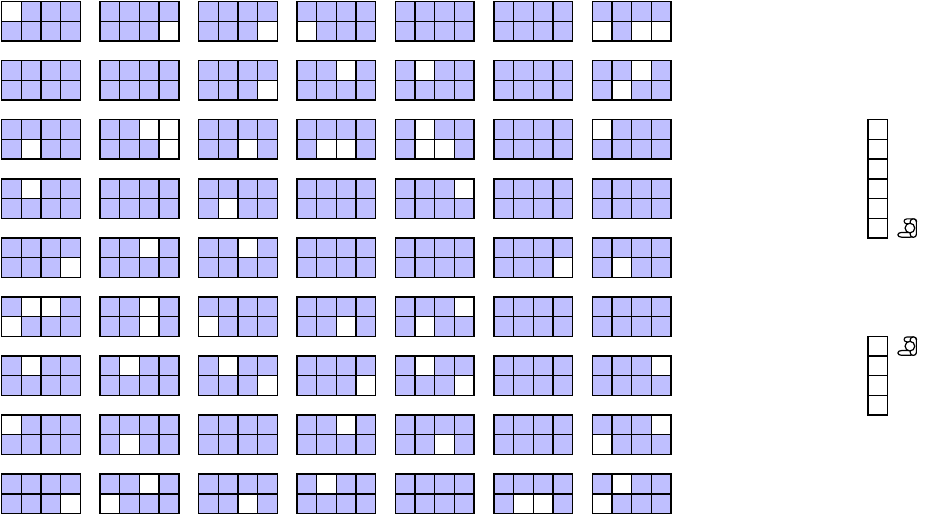}\caption{{\SystemFiveHundreds} with asymmetric {\outputStations}.\label{fig:504-layout}}
\end{figure}

For more realistic tests we create a larger system with 504 places
and 441 {\pods}. The total number of places at the {\outputStations}
is equal to the minimal number of robots we would expect in a warehouse
of this size. To make the problem harder we use asymmetric {\outputStations}.
See \Figref{504-layout}. The costs between places and {\outputStations}
correspond to distances that a real robot would travel. The initial
positions of the {\pods} at time $t=0$ are random.

Like in the small system, we use for the {\pod} weights the truncated
geometric distribution with weight $\podweight 1$ selected in such
a way that $\podweight 1/\podweight{441}=20$. Thus, the weight of
the most frequently used pod is $\podweight 1=0.0071399315$. For
the stations we assume that station $1$ works faster than station
$2$ and we set their weights to $\stationweight 1:=0.6$ and $\stationweight 2:=0.4$.
(A higher station weight means more throughput through that station.)
The number of time steps is $20\,000$.

\section{Algorithms}\label{sec:algorithms}
In this section we will analyze several algorithms to optimize our
warehouse model: three simple, like cheapest place, fixed place and
random place; two classical solving approaches, like iterative BIP
and Genetic Algorithms; and finally we will create our own heuristic.
We will not only use the algorithms for optimization, but we will
also use them to better understand the underlying {\ProblemName}
Problem. At the end of each analysis we will briefly list the advantages
and disadvantages of each algorithm.

\subsection{Cheapest place\label{subsec:cheapest-place}}

As the name suggests, a cheapest-place algorithm selects for a current
{\pod} the cheapest available place in the storage. It completely
ignores how this decision will influence the costs of other {\pods}
in the system. In the real world, ``cheapest'' often refers to distance
or time. In the \emph{{\Rawsimo}} simulation (see \citet{Merschformann:2018})
this algorithm is called \emph{Nearest}.

Even for this very simple algorithm we have the first technical problem:
``What do \emph{costs} mean?'' Here we talk about abstract costs
-- not about their physical meaning like money, time, or distance.
We introduce three possibilities:
\begin{enumerate}
\item \emph{Cheapest to storage}. In this case we consider only the costs
of moving the current {\pod} from a {\outputStation} to the storage,
that is $\coststostation(\placeelement,\stationelement)$ only. We
ignore how much it will cost to move this {\pod} to a {\outputStation}
again. The NearestPod algorithm in {\Rawsimo} works in this way.
Sometimes it is reasonable to use these costs. This is when the fact
that it would be cheap to move a {\pod} into a place implies that
it would also be cheap to move it out of this place.
\item \emph{Cheapest on average}. In this case we assign to each place average
costs
\begin{equation}
\begin{aligned}\AverageCosts:\placeset & \rightarrow\mathbb{R}\\
\placeelement & \mapsto\sum_{\stationelement\in\stationset}\left(\coststostation(\placeelement,\stationelement)+\costsfromstation(\stationelement,\placeelement)\right)\cdot\StationRelativeFrequency{\stationelement}
\end{aligned}
\label{eq:average-costs}
\end{equation}
where $\StationRelativeFrequency{\stationelement}$ is the proportion
of {\pods} who leaves the storage area to station $\stationelement$
\[
\StationRelativeFrequency{\stationelement}:=\frac{1}{N}\sum_{i=0}^{N-1}\Indicator{\stationprojection{\nextpod}=\stationelement}.
\]
The average costs can be helpful if we want to ``globally rank''
the places -- independently from the {\outputStations} and from
the {\pods}. We will use these costs for a fixed-place approximation
and for the genetic algorithm later.
\item \emph{Cheapest} \emph{decision}. In these costs we use the known future
station $\stationelement_{\text{to}}$, which the {\pod} will move
later to. The special value $\stationelement_{\text{to}}=0$ means
that the {\pod} will not leave the storage
\begin{equation}
\begin{aligned}\DeterministicCosts:\placeset\times\stationset\times(\stationset\cup\{0\}) & \rightarrow\mathbb{R}\\
(\placeelement,\stationelement_{\text{from}},\stationelement_{\text{to}}) & \mapsto\coststostation(\placeelement,\stationelement_{\text{from}})+\costsfromstation(\stationelement,\stationelement_{\text{to}})\cdot\Indicator{\stationelement_{\text{to}}\neq0}.
\end{aligned}
\label{eq:deterministic-costs}
\end{equation}
To keep the cheapest-decision costs simple, we ignore the future costs
caused by the terminal costs $C_{0}$.
\end{enumerate}

\subsubsection{Properties of cheapest-place algorithms}

We analyze the storage-area chart created by a cheapest-place algorithm.
In our small system, it does not matter which of the three costs we
use to determine the cheapest place: the result is the same.

\begin{figure}[h]
\begin{centering}
\includegraphics{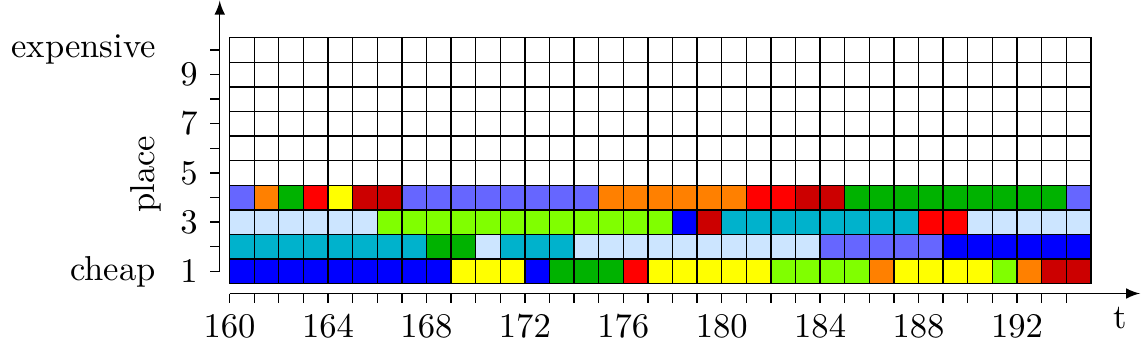}
\par\end{centering}
\caption{Cheapest-place algorithm. Changing of the storage area during the
time span $[160,196)$. (From dark blue to dark red: least used to
most used {\pods}.)\label{fig:cheapest-place-storage-chart}}
\end{figure}

A cheapest-place algorithm generates a distinct pattern in storage-area
charts -- it uses as few places as possible. See \Figref{cheapest-place-storage-chart}.
The algorithm appears to be optimal -- but it is not. The problem
is that it often puts rarely used {\pods} close to the output stations.
There, these {\pods} waste resources for a long time.

The final question we want to answer in this subsection is: ``Is
the cheapest-place algorithm optimal?'' Our experiments suggest that,
in general, it is not. But under special conditions -- uniform usage
of {\pods} -- it provides pretty good results. To be more precise,
our experiments show that ``pretty good results'' actually mean
that the cheapest-place algorithm provides similar results for different
distributions of pods, but under the special condition -- uniform
usage of {\pods} -- the other algorithms become worse.
\begin{rem}
A cheapest-place solution is closer to an optimal solution the more
uniformly the {\pods} are used: see \Figref{optimality-the-cheapest-place}.
There we have tested cheapest-place algorithm on the small test system
with different degrees of uniformity:
\begin{itemize}
\item \emph{random\ geometric} The departures of the {\pods} are random
and \emph{not} uniformly distributed. We use geometric weights from
the small test system as described in \Subsecref{test-system-10}.
The stations are randomly selected.
\item \emph{random\ uniform} The departures are random and uniformly distributed.
We use equal weights for {\pod} selection. The stations are randomly
selected.
\item \emph{periodic\ random} The departures are blocks of random permutations
of {\pods}. For example (1, 4, 5, 6, 7, 3, 9, 2, 8, 10), (4, 8, 7,
1, 3, 2, 4, 5, 6, 9, 10),\ldots{} Each permutation is uniformly distributed.
If a {\pod}, that needs to depart according to the generated sequence,
is not in the storage area, we use some simple correction. The stations
are randomly selected.
\item \emph{periodic} The departures are deterministic and uniformly distributed.
The departures are repeating sequences (1, 2, 3, 4, 5, 6, 7, 8, 9,
10) and the destination stations are repeating sequences (1, 2).
\end{itemize}
With periodic departures, the cheapest-place solution is optimal.

\begin{figure}
\begin{centering}
\includegraphics{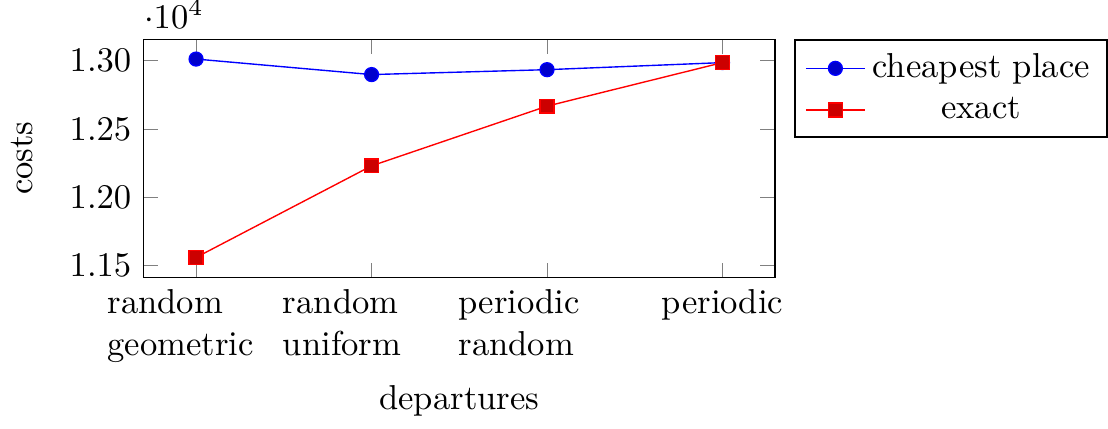}
\par\end{centering}
\caption{Optimality of the cheapest place algorithm for {pod} departures with
different degree of uniform usage.\label{fig:optimality-the-cheapest-place}}
\end{figure}
\end{rem}

\subsubsection{Advantages and disadvantages}

A cheapest-place algorithm has several advantages and disadvantages:
\begin{itemize}
\item[$\oplus$]  It is simple. The cheapest-to-storage version of the algorithm can
be easily applied for more complex simulations without modification.
\item[$\oplus$] It appears to work well, when the {\pods} are used similarly often. 
\item[$\oplus$] It appears to provide good results in the {\Rawsimo} simulations
if combined with optimization policies for other RMFS decision problems,
see \citet*[Section 8]{merschformann-lamballais-de-koster-suhl:2018}.
\item[$\ominus$] It does not work well when the {\pods} are used with very different
frequencies.
\item[$\ominus$] If the {\pods} are used with different frequencies and they are
already optimally ordered in the storage area, the cheapest-place
algorithm will destroy this order and proceed to work not optimally.
\end{itemize}

To evaluate the quality of our algorithms we need some reference algorithms.
The best reference is the exact solution, but unfortunately it is
only available for small and simplified models. More practical reference
algorithms are: fixed place and random place.

\subsection{Fixed place}

A fixed-place algorithm always assigns the same place to the same
{\pod}. That means we look for an optimal function
\[
\begin{aligned}\FixPlaces:\podset & \longrightarrow\placeset\\
\podelement & \mapsto\FixPlaces(\podelement).
\end{aligned}
\]
which assigns a place to every pod $h$ which will leave the {\outputStations}.
We calculate an optimal function $\FixPlaces$ with BIP\@. To prevent
conflicts, we assume that in the initial state all the {\pods} in
the storage area have been assigned by $\FixPlaces$ too.

For BIP, we represent the function $\FixPlaces$ with decision variables
$x_{\podelement\placeelement}\in\{0,1\}$. $x_{\podelement\placeelement}=1$
means ``assign {\pod} $\podelement$ to place $\placeelement$,''
or more formally: $x_{\podelement\placeelement}=1\Longleftrightarrow\FixPlaces(\podelement):=\placeelement$.
From departures $\nextpods$, we count how frequently each {\pod}
$\podelement$ visits station $\stationelement$ and how frequently
it comes back. We store this information in $\ToStationCount(\podelement,\stationelement)$
and $\FromStationCount(\podelement,\stationelement)$. Now we can
define costs $c_{\podelement\placeelement}$ if {\pod} $\podelement$
will always return to place $\placeelement$
\[
c_{\podelement\placeelement}:=\sum_{\stationelement\in\stationset}\left(\ToStationCount(\podelement,\stationelement)\cdot\coststostation(\placeelement,\stationelement_{\text{from}})+\FromStationCount(\podelement,\stationelement)\cdot\costsfromstation(\stationelement,\stationelement_{\text{to}})\right).
\]
We solve the minimization problem
\[
\min\sum_{\podelement\in\podset}\sum_{\placeelement\in\placeset}c_{\podelement\placeelement}x_{\podelement\placeelement}
\]
subject to constraints: assign every pod
\begin{equation}
\sum_{\placeelement\in\placeset}x_{\podelement\placeelement}=1,\qquad\forall\podelement\in\podset,\label{eq:only-one-place}
\end{equation}
and no more than one pod per place
\begin{equation}
\sum_{\podelement\in\podset}x_{\podelement\placeelement}\leq1,\qquad\forall\placeelement\in\placeset.\label{eq:eq:no-more-than-one-pod-per-place}
\end{equation}

\begin{rem}
Determining of the optimal places for the fixed-place algorithm is
an assignment problem. The {\pods} are tasks and the places are the
agents. When we assume that the {\pods} go with the same relative
frequency to the {\outputStations} we can find very fast solution
with following algorithm: Sort {\pods} by their usage frequency and
write them to a list. Sort places by their average costs (\ref{eq:average-costs})
and write them to another list. Assign the $i$-th {\pod} to the
$i$-th place in the corresponding lists.
\end{rem}
\begin{rem}
The fixed-place algorithm is an extreme case of zoning strategy. Zoning
means, we divide the storage area into different zones. {\Pods} with
different frequencies of usage go to different zones: see \citet*{lamballais-roy-de-koster:2017}.
The fixed-place algorithm has as many zones as there are {\pods}
in the system.
\end{rem}

\subsubsection{Properties of fixed-place algorithms}

\begin{figure}[h]
\begin{centering}
\includegraphics{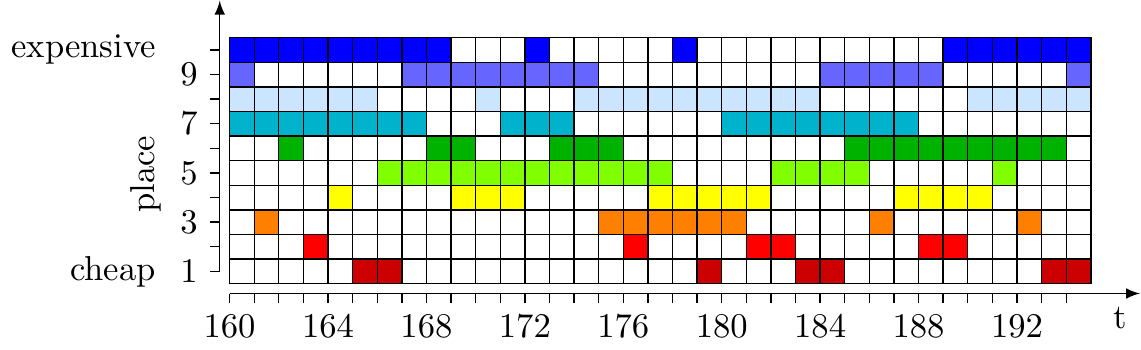}
\par\end{centering}
\caption{Fixed-place algorithm. Changing of the storage area during the time
span $[160,201)$.\label{fig:fixed-place-chart}}
\end{figure}

We analyze a storage-area chart of a system with a fixed-place policy.
See \Figref{fixed-place-chart}. The typical pattern of the fixed-place
algorithm is that there is only one non-white color in every line.
We see that the algorithm does not use a lot of cheap places. This
means that the fixed-place algorithm is in general not optimal. The
effect of unused places is very clear in the small test system, because
most of the {\pods} are at the output stations. In contrast, in a
more realistic system, where there are many more {\pods} in the storage
area than in the {\outputStations}, we expect the fixed-place algorithm
to be pretty good.

When we know the future departures for a long period of time, we can
use the fixed-place algorithm as a good upper bound for the long-term
cost. Here we assume, that we need only a small fraction of time and
costs to bring a system from some arbitrary initial state, where {\pods}
are not optimally distributed, to an optimal state, where places are
assigned by $\FixPlaces$. Or if we assume that we can actively reposition
the {\pods} before applying the algorithm, such as in \citet{Merschformann:2018},
for a small fraction of the total costs. Unfortunately, this does
not work for many real-world settings, because we cannot predict future
departure of the {\pods} for a long period of time. On one day, we
start to move the system to some optimal state, but on the next day
the customers order different things and we need to move to another
optimal state. In this case, the fixed-place algorithm becomes a good
lower bound for the optimal costs.

\subsubsection{Advantages and disadvantages}

Here are advantages and disadvantages of the fixed-place algorithm:
\begin{itemize}
\item[$\oplus$] It is simple.
\item[$\oplus$] We expect that it works well for larger systems.
\item[$\ominus$] It requires an (expensive) rearrangement of {\pods} before we can
use the algorithm.
\item[$\ominus$] It is not practical if we have seasonal changes in {\pod} usage.
\end{itemize}

\subsection{Random place}

We use the random place assignment as a performance reference for
other algorithms. In this subsection \emph{random} means: every time,
we select one place from all available free places with equal probability
independently, from our previous decisions\footnote{The only dependence from the previous decision is that the previous
decision changes the set of free places.}. From a modeling point of view the random strategy stands for ``we
don't know'' and for ``we don't care.'' It shows what happens if
we do not optimize. The performance of the random algorithm is usually
bad, but in some cases stochastic magic happens and random becomes
optimal. 

\begin{figure}[h]
\centering{}\includegraphics{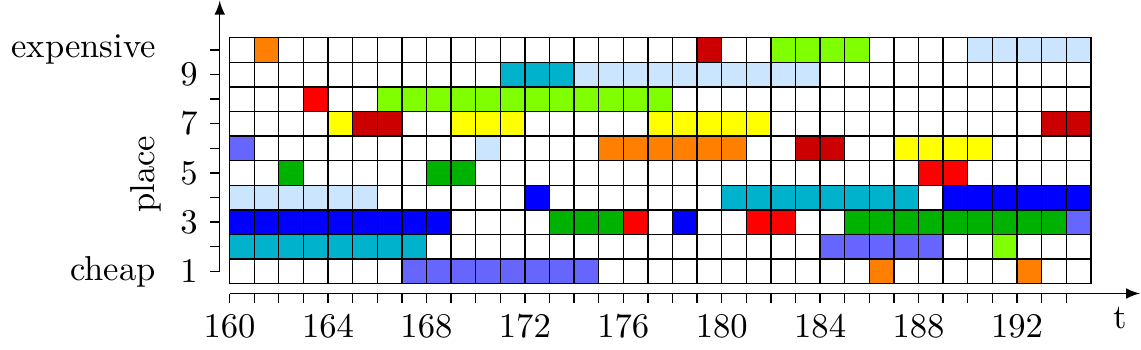}\caption{Random algorithm. Changing of the storage area in time.\label{fig:evo-random-chart}}
\end{figure}

For the sake of completeness, we provide a storage-area  chart of
the random algorithm. See \Figref{evo-random-chart}. It shows, how
the resources are wasted: The algorithm does not use places close
to the output stations efficiently. It puts frequently used {\pods}
too far away.

Finally, here is a brief list of advantages and disadvantages of the
random algorithm:
\begin{itemize}
\item[$\oplus$]  It is simple.
\item[$\oplus$]  It is a good reference for performance comparison between different
algorithms.
\item[$\ominus$]  It provides bad results.
\end{itemize}
\bigskip{}

In \Subsecref{binary-integer-programming} we described how to solve
the {\ProblemName} Problem exactly with BIP\@. This approach works
well for the small test system but fails for more realistic systems,
like for example the \systemFiveHundreds. The reason for the failure
is a lot of constraints (4\,158\,636\,886), about $|\placeset|\cdot|\podset|\cdot|\timespace|\cdot=504\cdot441\cdot20\,000$,
and possible NP-hardness of the {\ProblemName} Problem. In the next
subsection we will use three heuristic methods, which should still
provide good results.

\subsection{Iterative binary integer programming\label{subsec:Iterative-binary-integer}}

A calculation of an optimal solution is computationally intensive.
Instead of calculating the optimal solution for the entire time $\timespace$,
we split the time into $\NumTimeIntervals$ intervals $I_{k}=[t_{k},t_{k+1})$
with $t_{0}=0$ and $t_{\NumTimeIntervals+1}=N$. We call the set
of these intervals $\mathcal{I}$. Then we minimize (\ref{eq:bin-objective})
restricted to an interval
\[
\sum_{t\in I_{k}}\sum_{\placeelement\in P}\left(\costsfromstation(\FromStation t,\placeelement)+\coststostation(\placeelement,\ToStation t)\cdot\indicator{\ToStation t=0}\right)\cdot x_{t\placeelement}\qquad\forall I_{k}\in\mathcal{\SetOfTimeIntervals}
\]
one by one.

The conditions for the initially busy places (\ref{eq:initial-conditions-busy-constraints})
become 

\[
\InitialBusyEnd p(t_{k})\leq\underbrace{\BusyStart t}_{=t+1}x_{t\placeelement}+M_{\text{big}}(1-x_{t\placeelement})\qquad\forall t\in\timespace\setminus\{0\},\placeelement\in\placeset,
\]
where the end of the busy periods $\InitialBusyEnd p(t_{k})$ is defined
as

$\InitialBusyEnd p(0):=\InitialBusyEnd p$ 
\[
\InitialBusyEnd p(t):=\max\{\InitialBusyEnd p(t-1),\BusyEnd{t-1}(t-1)x_{t-1,\placeelement}\}.
\]

The conditions for previously busy places (\ref{eq:previous-decisions-busy-constraints})
become

\[
B_{\tau}x_{\tau p}\leq\underbrace{\BusyStart t}_{=t+1}\cdot x_{t\placeelement}+M_{\text{big},k}(1-x_{t\placeelement})\qquad\forall t\in I_{k},\tau\in I_{k},\tau<t,\placeelement\in\placeset
\]
with $M_{\text{big},k}=\max(I_{k})+1$.
\begin{rem}
\label{rem:iterative-bip}One may see the iterative BIP as a sequence
of solving of small {\ProblemName} Problems for each time interval
$I_{k}$. In each small problem, the terminal costs consist of the
future costs for placing {\pods} which will leave the storage area
outside of $I_{k}$. The final state becomes the initial state of
the subsequent problem on $I_{k+1}$. The fact, that we ignore costs
from the previous decisions on $I_{k-1}$, does not change the solution
on $I_{k}$, because these costs are only a constant subtracted from
the total costs on $I_{k}$. 
\end{rem}
The main advantages and disadvantages of the iterative BIP method
are:
\begin{itemize}
\item[$\oplus$]  It provides good results.
\item[$\ominus$]  It is still computationally intensive.
\item[$\ominus$]  It is not flexible.
\end{itemize}

\subsection{Genetic algorithms}

A solution for a deterministic warehouse problem is a sequence of
actions $\actiont 0{N-1}$ which optimally assigns resources (places)
under complex conditions. This kind of combinatorial problem suggests
the use of genetic algorithms. 

The goal of this section is not to find the best genetic algorithm
or the best set of parameters. It is rather an attempt to find out
how one may use genetic algorithms for the {\ProblemName} Problem,
what problems may occur, and what we can learn from the first results. 

We implemented genetic solvers with a python package,\ DEAP:\ see
\citet*{deap:2012}.

\subsubsection{First approach\label{subsec:evo-first-approach}}

For a genetic algorithm we must represent our solution as a chromosome.
The first natural attempt is to use as the chromosome a sequence of
actions. That means a sequence of places $\placeelement\in\placeset$
and zeros. As you may recall, a zero action means: ``do not send
any {\pod} to storage.'' In this paper, we will call this algorithm
\emph{genetic 1}.

Every mutation of a chromosome is a random change of some action.
The probability that a chromosome element will mutate is $3/N$, where
$N$ is the total number of actions. This leads to three changes in
the entire chromosome on average. We do not want to have too many
changes, because they easily create an unfeasible solution. We also
do not want to have too few changes, because then the improvement
rate sinks. If an action randomly changes, it changes to a uniformly
distributed value $\placeelement\in\placeset$. We use a random two-point
cross-over.

As an initial solution we use a random feasible solution. We tried
to use the cheapest-place solution, but our genetic algorithm could
not improve it. The fitness value of a chromosome is the average costs
per time step. 

\begin{figure}[h]
\begin{centering}
\includegraphics{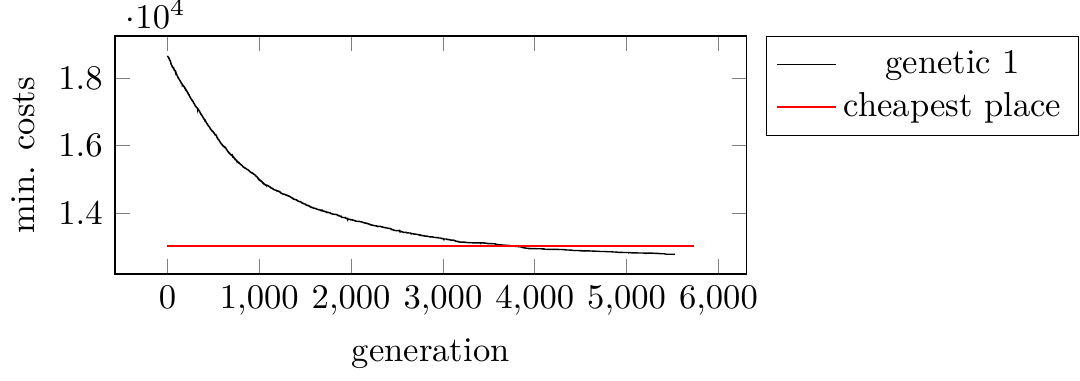}
\par\end{centering}
\caption{Minimal costs by a genetic algorithm, whose chromosome is places.
See \Subsecref{evo-first-approach}.\label{fig:genetic-1-intermediate-results}}
\end{figure}

We ran the algorithm until the last 100 generations could not improve
the best results of the previous generations. The population size
per generation was 100 individuals. In our small test system, the
algorithm converged very slowly: only after 3761 generations did it
become better than the cheapest-place algorithm. See \Figref{genetic-1-intermediate-results}.

\subsubsection{Second approach\label{subsec:evo-ord-second-approach}}

\global\long\def\EvoOrder{\gamma}
\global\long\def\EvoOrderClose{\gamma_{\text{close}}}
\global\long\def\EvoOrderFar{\gamma_{\text{far}}}
\global\long\def\EvoOrderZigzag{\gamma_{\text{zigzag}}}
\global\long\def\ChromosomeClose{\chi_{\text{close}}}
\global\long\def\ChromosomeFar{\chi_{\text{far}}}
\global\long\def\ChromosomeZigzag{\chi_{\text{zigzag}}}

\global\long\def\ChromosomeMutatedClose{\chi_{\text{mutated, close}}}
\global\long\def\ChromosomeMutatedFar{\chi_{\text{mutated, far}}}
\global\long\def\ChromosomeMutatedZigzag{\chi_{\text{mutated, zigzag}}}

In our first implementation we had a chromosome made of places, this
caused a lot of not feasible solutions (ca. 23 \% for the small test
system). The reason for this was that the algorithm randomly chose
a place but did not care whether this place was free. We improve this
by constructing a chromosome which considers only \emph{free} places.
In this paper, we will call this algorithm \emph{genetic }2.

For this new algorithm, we need a particular order of places $\EvoOrder$.
That means, a bijective function
\begin{align*}
\EvoOrder:\{0,\ldots,|\placeset|-1\} & \longrightarrow P\\
\placeelement & \mapsto\gamma(\placeelement).
\end{align*}

Every chromosome element for time $t$ is an index from 0 to $|D(\sysstate(t))|-1$.
It shows which free place from $D(\sysstate(t))$ -- sorted by $\gamma$
-- to choose. The element at the beginning has index $0$. The following
example demonstrates how the new algorithm works.

\begin{figure}[H]
\begin{centering}
\subfloat[System at time $t=0$.\label{fig:evo-at-t-0}]{\centering{}\includegraphics{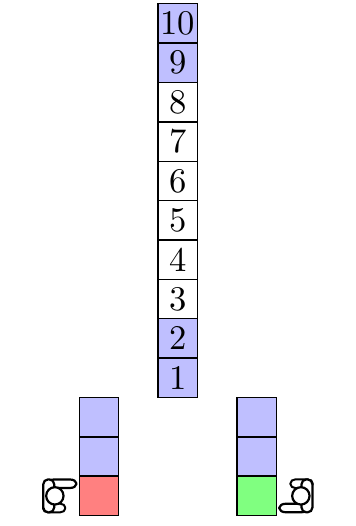}}\quad{}\subfloat[System at time $t=1$.\label{fig:evo-at-t-1}]{\centering{}\includegraphics{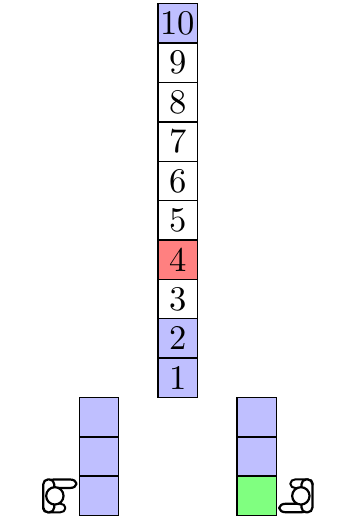}}\quad{}\subfloat[System at time $t=2$.\label{fig:evo-at-t-2}]{\centering{}\includegraphics{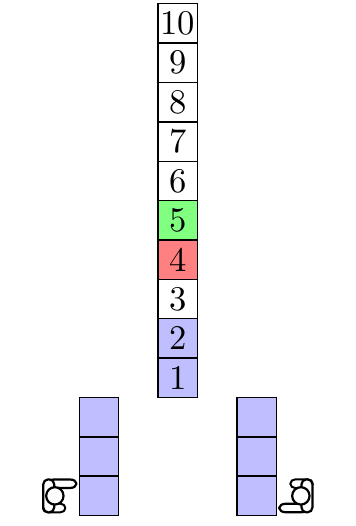}}
\par\end{centering}
\begin{centering}
\subfloat[Mutated system at time $t=1$.\label{fig:evo-mutated-t-1}]{\centering{}\includegraphics{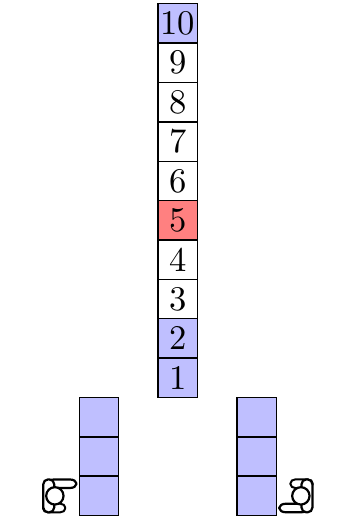}}\quad{}\subfloat[Mutated system with places ordered by the distance -- $\protect\EvoOrderClose$
and $\protect\EvoOrderFar$ -- at time $t=2$.\label{fig:evo-mutated-distance-order-t-2}]{\centering{}\includegraphics{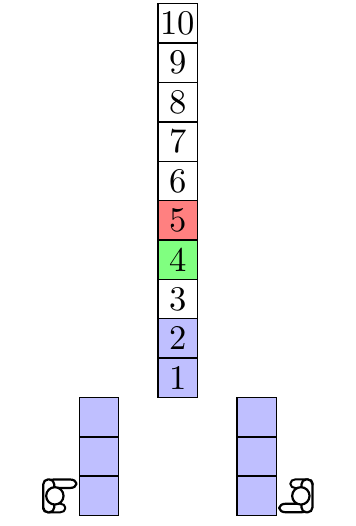}}\quad{}\subfloat[Mutated system with  order $\protect\EvoOrderZigzag$ at time $t=2$.\label{fig:evo-mutated-zigzag-t-2}]{\centering{}\includegraphics{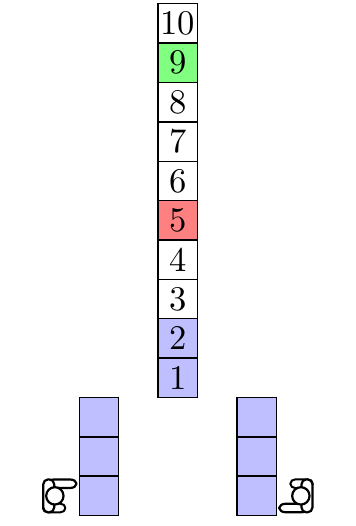}}
\par\end{centering}
\caption{System states when using genetic algorithm.}
\end{figure}

\begin{example}
Consider a chromosome $\ChromosomeClose:=(1,1)$. We order the places
by their distances to the stations. The closest one is in the front
\begin{equation}
\EvoOrderClose:=\zerobasedlist{1,2,3,4,5,6,7,8,9,10}.\label{eq:evo-chromosom-close}
\end{equation}
At time $t=0$ the system has a state shown in \Figref{evo-at-t-0}.
As you may recall, the decision to occupy a free place at time $t$
is made at time $t-1$. At time $t=1$ a {\pod} at place $9$ will
arrive at the left queue. The sequence of allowed actions ordered
by $\gamma_{\text{close}}$ is then $D(\sysstate(0))=\zerobasedlist{\underbrace{3}_{\mathllap{\text{0th place}}},\underbrace{4}_{\mathrlap{\text{ 1st place}}},5,6,7,8,9}$.
The red {\pod} goes to place $4$. The new state is shown in \Figref{evo-at-t-1}.

At time $t=2$ a {\pod} from place $10$ will arrive at the right
queue. The ordered sequence of allowed actions at time $t=1$ is $D(\sysstate(1))=\zerobasedlist{3,\underbrace{5}_{\mathclap{\text{1st place}}},6,7,8,9,10}$.
Following the commands in the chromosome, the algorithm selects place
$5$. The system changes to a new state which is shown in \Figref{evo-at-t-2}.
\end{example}

\subsubsection{Properties of genetic algorithms}

\begin{figure}
\centering{}\includegraphics{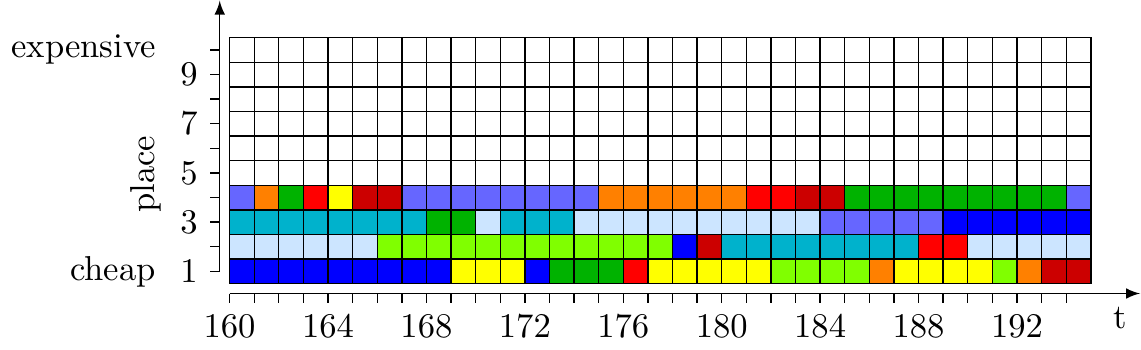}\caption{Genetic algorithm. Changing of the storage area during the time span
$[160,196)$.\label{fig:evo-ord-chart}}
\end{figure}

We focus on analysis of the improved genetic algorithm from \Subsecref{evo-ord-second-approach}.
As we can see in \Figref{evo-ord-chart}, within the plotted time
span, the algorithm puts a rarely used {\pod} in the far end of the
storage onto place $9$. It keeps other {\pods} close to the output
station and uses the closest place -- place 1 -- very intensively
for different {\pods}.

\paragraph{Place order $\gamma$ matters}

In our algorithm, a mutation is a uniformly distributed variable.
It seems that the place order $\gamma$ does not matter. And this
is true for a single mutated element of the chromosome (a permutation
of uniformly distributed variables has the same distribution). But
a chain effect on the subsequent actions depends a lot on the place
order $\gamma$. The following example demonstrates this.

\begin{figure}
\begin{centering}
\subfloat[Evolution of costs during optimization.]{\centering{}\includegraphics{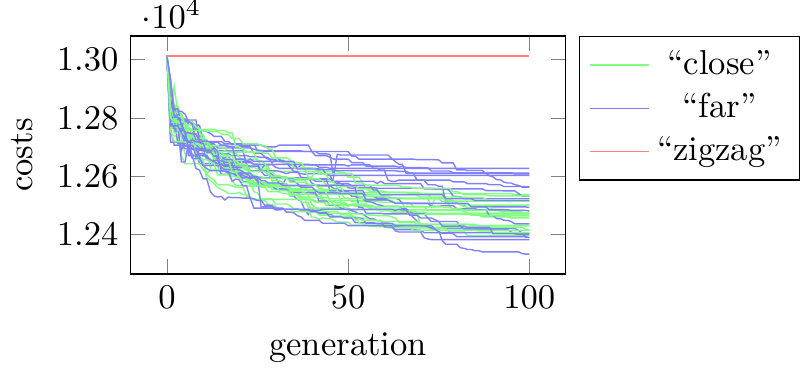}}\subfloat[Optimal results for every run.]{\centering{}\includegraphics{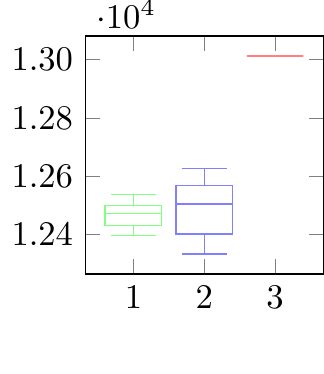}}
\par\end{centering}
\caption{Effect of different place orders on geometric algorithms.\label{fig:different-place-orders} }
\end{figure}

\begin{example}
\label{examp:order-matter}Consider three systems. All of them have
the same initial state and the same dynamics at times $0$, $1$ and
$2$ as shown in \Figref[s]{evo-at-t-0,evo-at-t-1,evo-at-t-2}. The
orders of the systems are:
\begin{align*}
\EvoOrderClose & :=\zerobasedlist{1,2,3,4,5,6,7,8,9,10},\\
\EvoOrderFar & :=\zerobasedlist{10,9,8,7,6,5,4,3,2,1},\\
\EvoOrderZigzag & :=\zerobasedlist{1,6,2,7,3,8,4,9,5,10}.
\end{align*}
The system changes following chromosomes
\[
\ChromosomeClose:=(1,1),\:\ChromosomeFar:=(5,5),\:\text{and \ensuremath{\ChromosomeZigzag}}:=(4,5).
\]

While the dynamics of the ``close'' and ``far'' systems are more
or less clear, the behavior of the ``zigzag'' system is less obvious.
We explain it step by step: At time $t=0$ the sequence of allowed
actions, ordered by $\EvoOrderZigzag$, is $\zerobasedlist{6,7,3,8,4,9,5}$.
The system chooses the fourth element -- place $4$. At time $t=1$
the ordered sequence of actions is $\zerobasedlist{6,7,3,8,9,5,10}$.
The system chooses the fifth element -- place $5$.

Now the first element of the chromosome mutates and the red {\pod}
moves at time $t=1$ onto place 5. See \Figref{evo-mutated-t-1}.
This corresponds to the new mutated chromosomes $\ChromosomeClose^{\text{mut}}:=(2,1)$,
$\ChromosomeFar^{\text{mut}}:=(5,5)$, and $\ChromosomeZigzag^{\text{mut}}:=(6,5)$.
The meaning of the second element in $\ChromosomeClose$, $\ChromosomeFar$
$\text{and \ensuremath{\ChromosomeZigzag}}$ changes. In $\ChromosomeClose$
and $\ChromosomeFar$ the values $1$ and $5$ now mean: ``move green
{\pod} to place 4.''; see \Figref{evo-mutated-distance-order-t-2}.
This mutated state at time $t=2$ is not very different from the non-mutated
state in \Figref{evo-at-t-2}. But for the system ordered by $\EvoOrderZigzag$
the change is more significant: The value $5$ in $\ChromosomeZigzag^{\text{mut}}$
now means ``move green {\pod} to place $9$'' (fifth free place
from $\zerobasedlist{6,7,3,8,4,9,10}$). The resulting state at time
$t=2$ on \Figref{evo-mutated-zigzag-t-2} differs much from the non-mutated
state in \Figref{evo-at-t-2}.
\end{example}
\Exampref{order-matter} shows how important the order $\EvoOrder$
is. We suggest selecting an order in such a way that the neighbor
places have similar costs. We made a numerical experiment with different
orders. We ran 20 optimizations for every order type, each with 100
individuals per generation for 100 generations. The results in \Figref{different-place-orders}
confirm the advantages of a ``smooth''\footnote{In this subsection the word \emph{smooth} means that a small change
in one decision does not cause a large change in the subsequent decision
and therefore total costs.} and disadvantages of a ``jumping'' order. To speed up the calculation
we sort the places by their average costs (\ref{eq:average-costs}).

\subsubsection{Advantages and disadvantages}

Here are advantages and disadvantages of the genetic algorithm. We
only consider the efficient version described in \Subsecref{evo-ord-second-approach}:
\begin{itemize}
\item[$\oplus$]  It is flexible.
\item[$\oplus$]  Decisions based only on free places are robust. When a place is
busy, the algorithm automatically resolves the problem by choosing
the next place.
\item[$\ominus$]  It is slow.
\end{itemize}

\subsection{Tetris\label{subsec:tetris}}

Every {\pod}-repositioning algorithm assigns a time interval and
a place to a {\pod}. We call these time intervals \emph{occupation
intervals}, because they show when the {\pod} will occupy a place
in the storage area. These algorithms cannot change the length and
the horizontal position of the occupation intervals and they must
ensure that the occupation intervals do not overlap.

\begin{figure}[H]
\centering{}\includegraphics{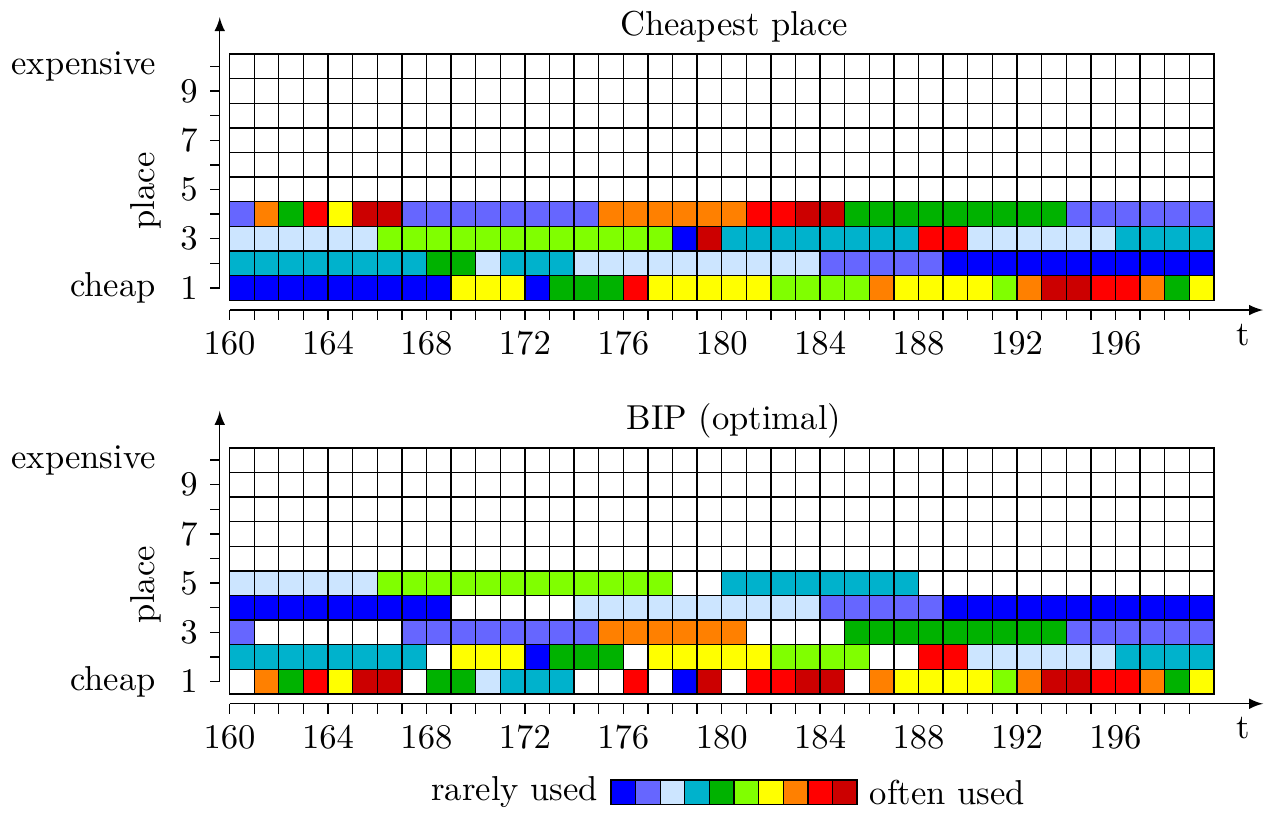}\caption{Comparison between two algorithms: cheapest-place and BIP\@. Both
algorithms are applied to the same instance of our small test problem.
{\Pods} are colored according their usage. Note, the cheapest-place
algorithm uses fewer places, but it is less optimal than the BIP\@.
This is because it is not important how many places an algorithm uses,
but rather how frequently it uses the cheapest places.\label{fig:colorful-comparison-charts}}
\end{figure}

\Figref{colorful-comparison-charts} explains this visually: It shows
two storage-area charts of two different algorithms. Each block represents
an occupation interval and the block's color refers to a particular
{\pod}. Both charts show that each algorithm moves blocks with the
same color up and down. The algorithms cannot move the blocks to the
left or to the right. They cannot split the blocks of the same color.
It cannot let the blocks overlap.

The visual comparison leads to a new algorithm. It is inspired by
a computer game, tetris.

\begin{figure}[H]
\begin{centering}
\subfloat[Create an initial feasible solution.\label{fig:tetris-explained-till-0}]{\begin{centering}
\includegraphics[scale=0.9]{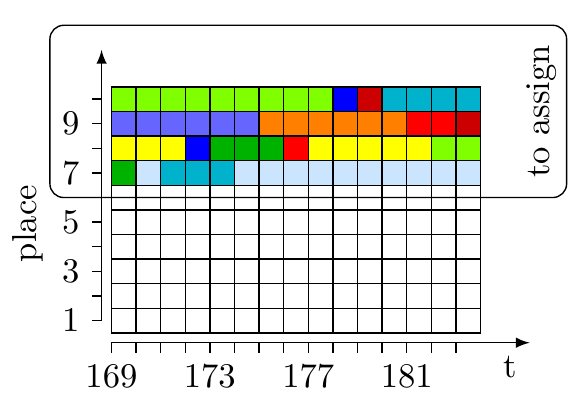}
\par\end{centering}
}\quad{}\subfloat[Assign the most frequently {\pod} first.\label{fig:tetris-explained-till-1}]{\begin{centering}
\includegraphics[scale=0.9]{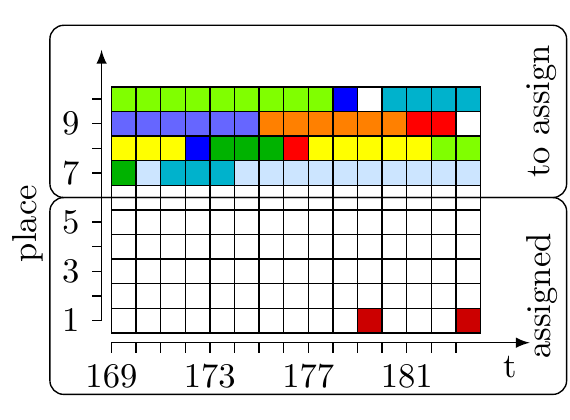}
\par\end{centering}
}
\par\end{centering}
\begin{centering}
\subfloat[Intermediate solution after assigning further three {\pods}: $2$,
$3$ and $4$.\label{fig:tetris-explained-till-4}]{\begin{centering}
\includegraphics[scale=0.9]{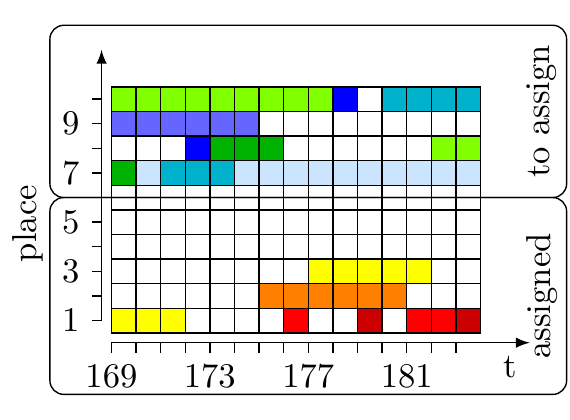}
\par\end{centering}
}\quad{}\subfloat[Result, after assignment of all {\pods} -- from $1$ to $10$.\label{fig:tetris-explained-till-10}]{\begin{centering}
\includegraphics[scale=0.9]{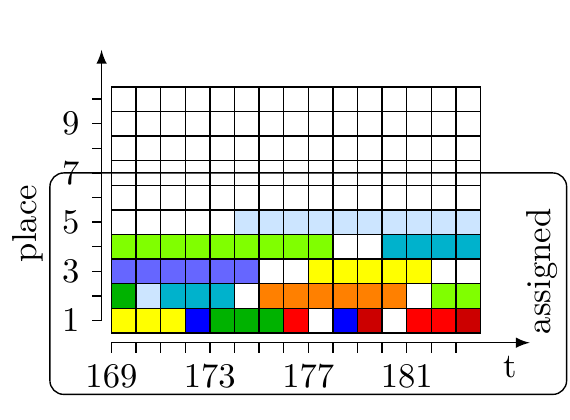}
\par\end{centering}
}
\par\end{centering}
\begin{centering}
\includegraphics[scale=0.9]{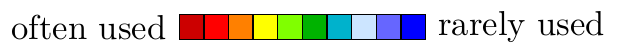}
\par\end{centering}
\centering{}\caption{Some steps from the tetris algorithm from \Subsecref{tetris}. Each
color shows how frequently a corresponding {\pod} is used during
the whole optimization time.\label{fig:tetris-explained}}
\end{figure}

\Figref{tetris-explained} explains the algorithm visually. In the
very first part of the algorithm, we create a feasible solution with
a lot of free cheap places. See \Figref{tetris-explained-till-0}.
The easiest way is to use a reverse version of the cheapest-place
algorithm -- the most-expensive-place algorithm. There, every time
a {\pod} leaves an output station, it will go to the most expensive
available place according to $\DeterministicCosts$ costs.

In the second part, we improve the initial solution: We select the
most frequently used {\pod} and move all its occupation intervals
to the cheapest free places. See \Figref{tetris-explained-till-1}.
If there is no cheaper place, we keep the old place. We start with
earlier occupation intervals and continue with later ones. We assign
the remaining {\pods} in the same way. See \Figref{tetris-explained-till-4,tetris-explained-till-10}.
A more formal pseudocode is in \Algoref{tetris}. 

In this paper, we call this algorithm \emph{tetris.} This single word
explains the idea quite well and is slightly shorter than its German
single-word alternative:\\
Regalhäufigkeitspriorisierungszukunftsverhaltenberücksichtigungsalgorithmus.

\begin{algorithm}[H]
\newcommand{\ArrivingShelf}{r}

\begin{algorithmic}[1]
	\Function{Tetris}{}
	\State  $\mathcal{I} \gets MostExpensivePlace()$ 

			$\mathcal{I} \subset \placeset \times \podset \times \timespace \times \timespace$
		
	\LonelyComment{Improve the initial solution.}
	\State  $\mathcal{I} \gets sort(\mathcal{I})$ 
		\Comment{\parbox[t]{50mm}{Sort the occupation intervals by 
	the pod frequency and then by the arrival times.}}
	\ForEach{$(\placeelement, \podelement, t_{\text{begin }}, t_{\text{end}}) \in \mathcal{I}$}
		\ForEach {$\placeelement^* \in \placeset$}
			\If{$costs(\placeelement^*)< costs(\placeelement)$ and 
				$p^*$ is free for $[t_{\text{begin }}, t_{\text{end}})$}
			\State
				$(\placeelement,  \podelement,t_{\text{begin }}, t_{\text{end}}) \gets
				(\placeelement^*,  \podelement, t_{\text{begin }}, t_{\text{end}})
				$
			\State
				$\action(t_{\text{begin }}-1) \gets p^*$
			\EndIf	
		\EndFor
	\EndFor
	\State \Return $\actiont{t}{\maxtime -1}$
	\Comment{\parbox[t]{50mm}{Return the sequence of actions.}}
	\EndFunction
\end{algorithmic}

\caption{\emph{Tetris} algorithm.\label{algo:tetris}}
\end{algorithm}

\subsubsection{Extension and variations of the algorithm}

\paragraph{Sort by occupation duration}

The first version of the algorithm relies on {\pod} frequencies during
the whole time. This makes it less suitable for seasonal changes of
{\pod} frequencies. To make it more robust against the seasonal changes,
we modify the second part of the algorithm: Instead of assigning the
most frequently used {\pods} first, we first assign {\pods} with
shorter occupation durations. According to our experiments, when we
do not have seasonal data, the algorithm based on the duration time
is a little less efficient than the frequency-based one. In contrast,
when we use seasonal data, the algorithm based on occupation duration
is better.

\begin{figure}[h]
\begin{centering}
\includegraphics{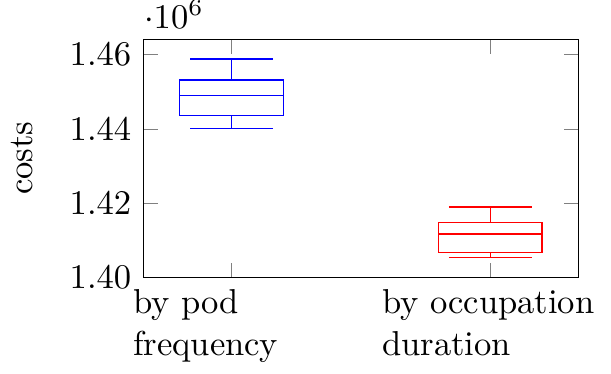}
\par\end{centering}
\centering{}\caption{Frequency-based tetris vs occupation-duration-based tetris applied
to seasonal data. Costs for a system with 504 places and 411 {\pods}
during the time $T=\{1,\ldots,10\,000\}$. Every $2000$ time steps,
the probability weights for the {\pods} change randomly. We made
20 tests with different random initial pod positions and different
{\pod} departures.\label{fig:tetris-seasoned-504}}
\end{figure}

\subsubsection{Properties of the tetris algorithm}

\begin{figure}[h]
\centering{}\includegraphics{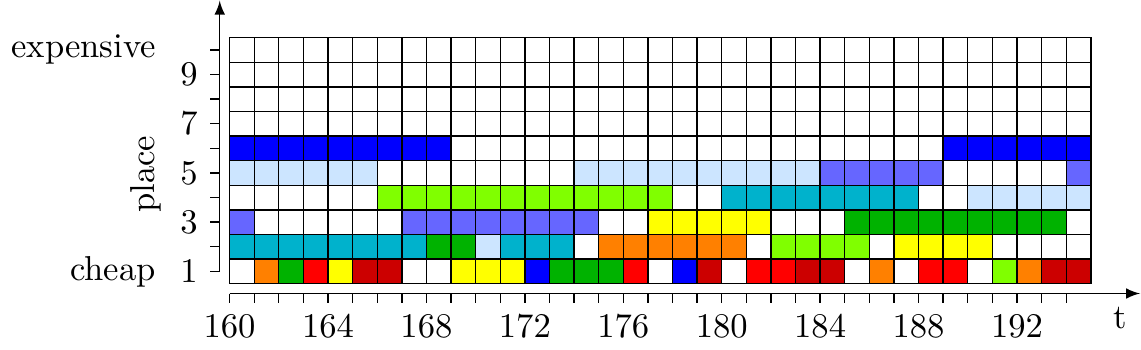}\caption{Tetris algorithm. Changing of the storage area during the time span
$[160,196)$.\label{fig:tetris-chart}}
\end{figure}

\Figref{tetris-chart} shows the storage-area chart produced by the
tetris algorithm. The algorithm puts frequently used {\pods} to the
front and {\pods} which spend a lot of time in the storage area to
the end of the storage. It tries to use the cheapest places even for
less frequently used {\pods} when a more frequent {\pod} cannot
use them. See for example the rarely used {\pod} during the time
span $[198,199)$ at place $1$ in \Figref{tetris-chart}. 

\subsubsection{Advantages and disadvantages}
\begin{itemize}
\item[$\oplus$] It appears to be good.
\item[$\oplus$] It is fast.
\item[$\oplus$] The version which is based on the length of occupation intervals
produces good results for seasonal {\pod} usage.
\item[$\ominus$] This algorithm is a heuristic which is made under some assumptions
about the usage frequency of the {\pods}. We do not know how well
this algorithm will perform for a particular real data set.
\end{itemize}

\section{Computational results}\label{sec:results}
We tested different algorithms on two types of system, the small one
from \Subsecref{test-system-10} and the {\systemFiveHundreds} from
\Subsecref{test-system-504}. For each system we used the same instance
for all tested algorithms. For BIP we used open-source library COIN-OR
CBC over the python PuLP interface and Gurobi. For fixed-place algorithm
we used COIN-OR CBC.

\begin{figure}
\begin{centering}
\subfloat[Total costs compared to costs of the random algorithm.]{\begin{centering}
\includegraphics{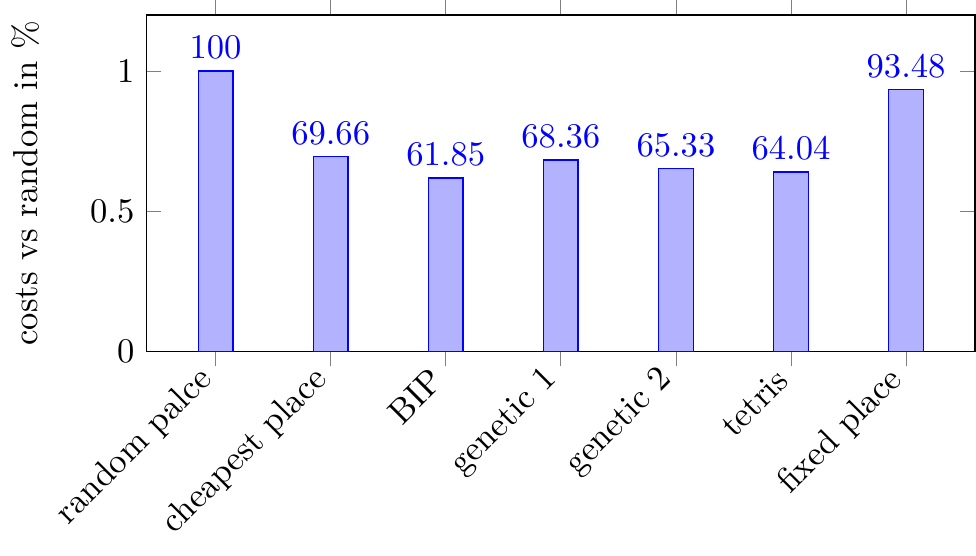}
\par\end{centering}
}
\par\end{centering}
\begin{centering}
\subfloat[Running times for all 1000 decisions in minutes. For BIP algorithm
we used Gurobi 8.01. We verified the BIP result with COIN-OR CBC with
running time 1.17 minutes.]{\centering{}\includegraphics{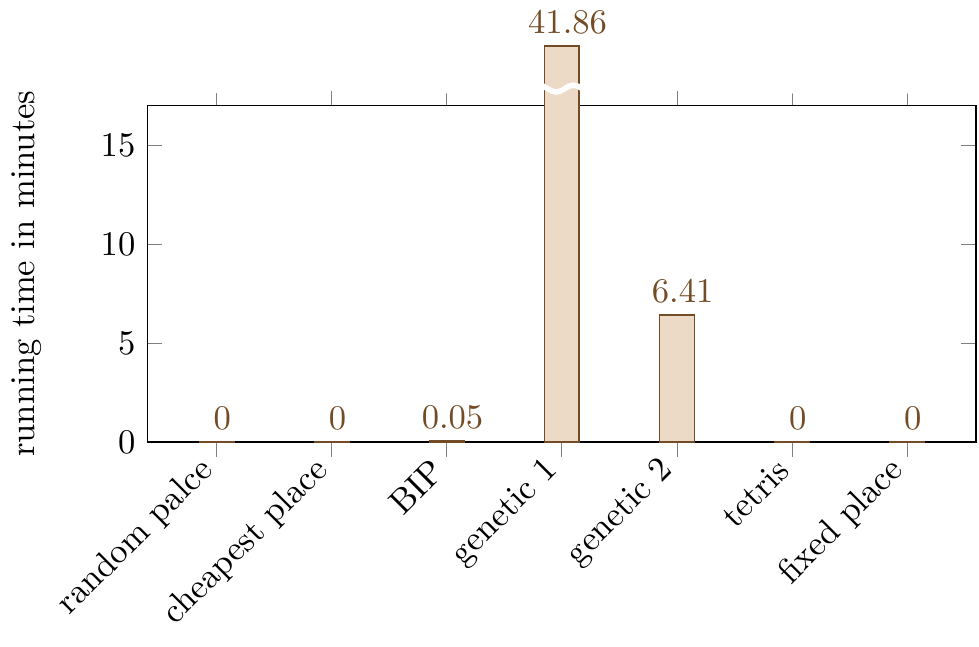}}
\par\end{centering}
\caption{Computation results for the small system with 10 places and 10 {\pods}
from \Subsecref{test-system-10}. \label{fig:results-for-10}}
\end{figure}

We use different algorithms for a small system with 10 places and
10 {\pods}. We ran the test on a notebook with an Intel Core i7-7600
CPU, 2.80 GHz, two cores (four logical cores), 16 GB RAM and Ubuntu
18.04. The results are in \Figref{results-for-10}. They show that
tetris provides pretty good results and it is also very fast.

\begin{figure}
\begin{centering}
\subfloat[Total costs compared to costs of the random algorithm.]{\begin{centering}
\includegraphics{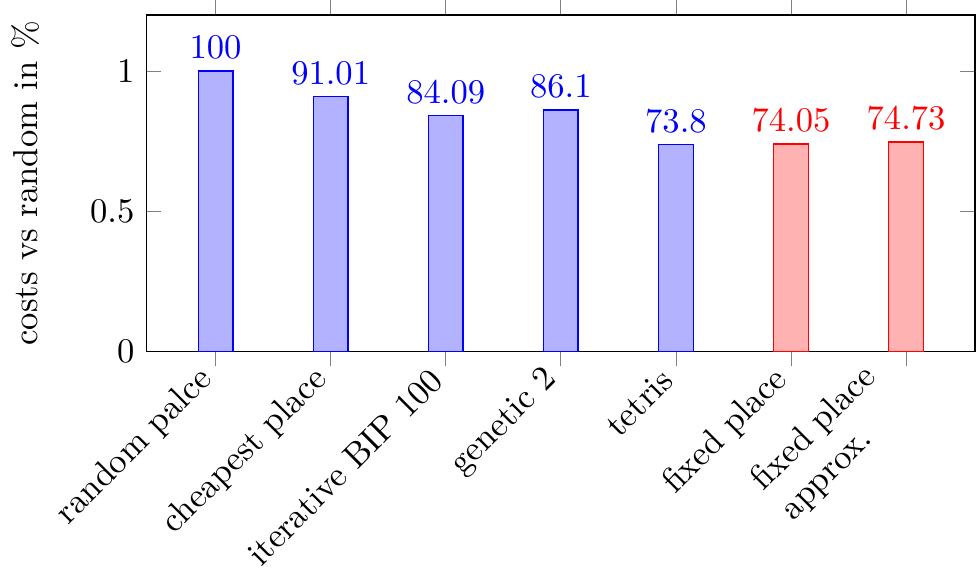}
\par\end{centering}
}
\par\end{centering}
\begin{centering}
\subfloat[Running times for all 20\,000 decisions in minutes.]{\centering{}\includegraphics{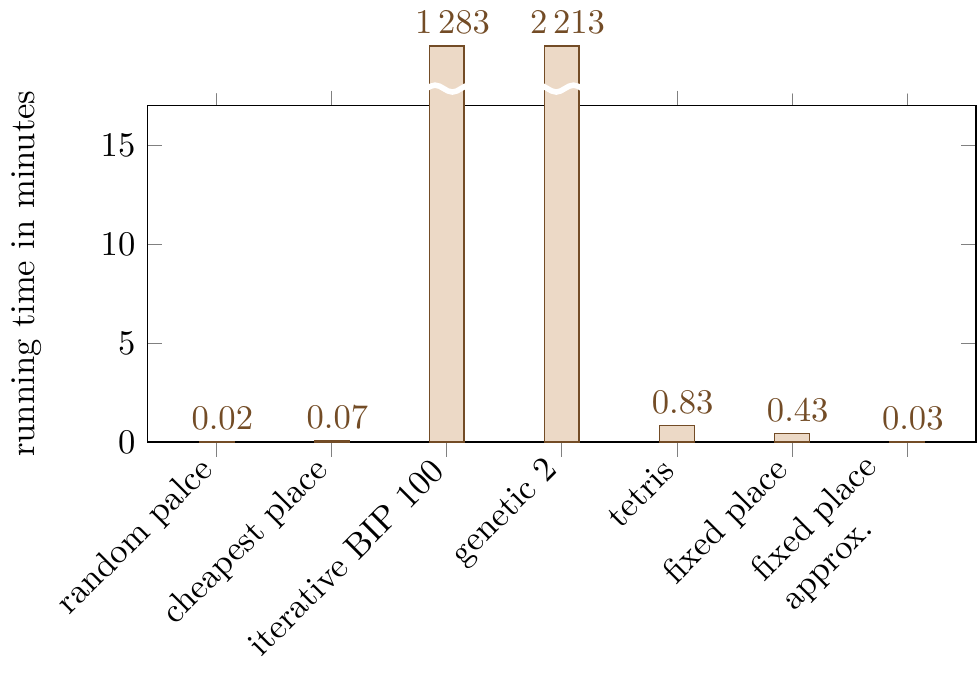}}
\par\end{centering}
\caption{Computation results for the {\systemFiveHundreds} with 504 places
and 441 {\pods} from \Subsecref{test-system-504}.\label{fig:results-for-504}}
\end{figure}

For a {\systemFiveHundreds} with 504 places and 441 {\pods} we used
only the random-place, cheapest-place, iterative BIP, genetic 2, tetris
and fixed-place algorithms. We did not use genetic 1 algorithm because
it demonstrates bad results for a small system. For computationally-intensive
genetic 2 algorithm we used a faster cluster with four Intel Xeon
E5-4670, 32 cores, 2.7GHz, with 64 GB RAM\@. For iterative BIP with
interval size $100$ we used a cluster with two Intel Xeon X5650,
16 cores, 2.67GHz, 32 GB RAM and Gurobi 7.5.2. Most of the time, the
iterative BIP solver spent for adding constraints. The results are
in \Figref{results-for-504}. They show that tetris provides a good
solution with 20 000 decisions within less than one minute.

Note, the fixed-place algorithms rearrange {\pods} of the system
with 504 \emph{before} starting the system. But we do not consider
any additional costs for this improvement of the initial state. This
corresponds to the \emph{active} {\pod} repositioning for free. In
contrast, all the other algorithms are not allowed to change the initial,
randomly generated, positions of the {\pods} before the system starts.
They must use only the \emph{passive} {\pod} repositioning. That
is why we cannot compare other algorithms with both fix-place algorithms
directly. Nevertheless we use the fixed-place results as a reference,
it shows what would happen if the system would starts with perfect
{\pod} positions and keep them the whole time.

\section{The problem with multiple stations}\label{sec:problem-with-multiple-stations}
The previous section demonstrates that it is important to know the
occupation time intervals of the {\pods} in the storage area. The
algorithms with the most optimal costs -- BIP and tetris -- use
this information. Unfortunately, in real life we do not have exact
information about the occupation time intervals. This is because the
picking times (and to a smaller extent the transport times) introduce
randomness to the system. Our occupation time intervals are only estimations
with estimation errors.

Although it is hard to determine the occupation time intervals exactly,
we do not really need it. What we really need is to know the order
in which {\pods} leave the storage area and the order in which {\pods}
return to the storage area. This information is partially available
in the real-world warehouses because they process customers' orders
in batches. For example, a warehouse collects customers' orders for
one hour without fulfilling them. Then it calculates an efficient
order in which the {\pods} will go to particular stations to process
the customers' orders. See for example the algorithms in \citet*{boysen-briskorn-emde:2017b}.

\paragraph{Single {\outputStation}}

For a single station, the information, about the order in which the
{\pods} will go to the {\outputStation} is fully sufficient for
most algorithms. For example, when {\pods} $A$, $B$, $C$ and $D$
leave the storage area in the order $ABCD$, they also arrive in the
same order, $ABCD$. No matter how fast or slow the picker is, the
order stays unchanged.

If we know the order of departures, we can expand them to \emph{rescaled}
occupation time intervals. The basic idea is as follows: Imagine you
record two warehouses simultaneously. One uses algorithm $A$ for
optimization and the other uses algorithm $B$. The pods depart in
both systems simultaneously and the picking times are the same. Algorithm
$A$ is better than algorithm $B$. Algorithm $A$ will still be better
than algorithm $B$ even if you play back your recording at different
speeds. Now you can speed up and slow down the playback to have exactly
one unit of time between the departures and between the arrivals;
these will make the occupation time intervals suitable for our algorithms.

To keep the things simple, we focus on the case when the queue at
the {\outputStation} is full: that means it contains $\stationmaxsize$
elements. Then, for the $i$-th {\pod} we use the time interval $\left[i+\stationmaxsize,j\right)$
as the occupation time interval in the storage area, where $j$ is
the index of the same {\pod} when it will depart from the storage
area next time. We give this information to one of our algorithms
and the algorithm returns the sequence of decisions. The precise time
of the {\pod} assignment does not matter, only its order.

When we know the order of the decisions, small timing problems may
still occur, but we can easily resolve them: It may happen that {\pod}
$A$ leaves the {\outputStation} too early and the assigned free
place is still occupied by another {\pod} $B$. To resolve this conflict,
the robot with {\pod} $A$ just waits until another robot removes
{\pod} $B$ according to the previously calculated sequence\footnote{In a rare case, when {\pod} $B$ requires the robot of {\pod} $A$,
we need more steps to exchange the {\pods}, but it is still fast
and physically possible. }. When {\pod} $A$ arrives at its place too late, the place can be
taken away. To prevent this, we reserve this place for $A$. We can
apply a similar conflict resolution to the {\outputStation}. When
the {\pod} comes too early, it must wait for other {\pods}. When
the {\pod} arrives too late, another {\pod} waits for it.

The conflict resolution requires additional time and therefore it
may have negative consequences for the throughput at the {\outputStation}.
To dampen these consequences we suggest having a queue with several
{\pods} at the output station. This queue acts like a buffer. It
dampens time fluctuations and reorganizes the {\pods} into the correct
order until they actually reach the picker.

\paragraph{Multiple {\outputStations}}

The situation becomes much more complicated, as soon as we have more
than one {\outputStation}. Also in this situation, the random fluctuations
in picking times and in transport times change the order in which
the {\pods} leave and arrive at the storage area. But the consequences
are now different. For example we may plan that {\pod} $A$ and then
{\pod} $B$ go to station $1$, and {\pod} $C$ and then {\pod}
$D$ go to station $2$. For our calculation we assume that the {\pods}
will leave the storage area in the order $ABCD$ and then return to
the storage area in the same order. But in reality the actual departure
order could be any of the following: $ABCD$, $ACBD$, $ACDB$, $CDAB$,
$CADB$ or $CABD$. We also can surely determine the order in which
the {\pods} will return. The rescaling of the time, like we did in
the single-station system, will not fix the departure and return orders.

If the fluctuations are not very high, then we still use a waiting
strategy to stick to the plan. To correct higher fluctuations we can
use long queues at the {\outputStations} or drift spaces. But if
we plan 1000 {\pod} assignments in advance, the difference between
the planned {\pod} order and the actual {\pod} order can become
too large. The output station will not be able to compensate time
differences and the whole system will slow down. Instead of waiting
we can use other strategies for larger timing errors:

\paragraph{Strategies for multiple {\outputStations}}

Often we will not be able to apply deterministic algorithms directly
in real-world settings: instead we can use them as concepts. For example,
the genetic algorithm, with its reaction on mutation and its focus
on free places, provides two possible strategies for {\pod} assignment:
\begin{enumerate}
\item Calculate the solution with estimated picking times. When a previously
assigned place is occupied in reality, then go to the next best (or
next worst) free place according to some order.
\item Calculate the solution with estimated picking times and translate
the solution into free-place indices, like the genome of the genetic
 algorithm in \Subsecref{evo-ord-second-approach}. For example, translate
``{\pod} $A$ goes to place $117$'' to ``{\pod} $A$ goes to
the $14$th free available place'' according to some preference list.
\end{enumerate}
We can derive another way to solve the problem from the tetris and
cheapest-place algorithms. We can look at the tetris algorithm as
a priority algorithm. In tetris, the most frequently used {\pods}
selects the best places first. That means these {\pods} have the
highest priority. Also, a {\pod} cannot select a place and occupy
it for a period of time if another {\pod} with higher priority wants
to have this place within this period of time. Also, the cheapest-place
algorithm is a sort of priority algorithm: a different one. It gives
to the current pod the highest priority to select any possible place.
Based on these priority concepts, we can create a new priority algorithm
which better suits the random occupation time intervals.
\begin{enumerate}[resume]
\item Estimate how frequently every {\pod} will be used. Assign a priority
to every pod based on its frequency. When a pod leaves a {\outputStation},
assign a free place to it based on its priority. That means: do not
take the best places if other {\pods} with higher priority want to
use them too. Apply the cheapest-place algorithm among {\pods} with
similar priorities.
\end{enumerate}

\section{Conclusion}\label{sec:conclusion}
We have shown that a very small mathematical model helps us to understand
essential parts of the Pod Repositioning Problem. It also provides
a fast heuristic -- the tetris algorithm. This algorithm appears
to be robust and flexible enough to be applied to larger instances.
We critically analyzed how implication from our mathematical models
can be or are already used in the real world. 

In \Secref{problem-with-multiple-stations}, we showed that for a
system with multiple {\outputStations}, a deterministic model is
not sufficient for real-world application. Unfortunately, to our knowledge,
there is no real warehouse with only one {\outputStation}. That means
that the deterministic model is only a starting point for a more realistic
stochastic model of a robotized warehouse. We also showed that in
a stochastic system we should focus on the estimation of the departure
and arrival order of the {\pods} from and into the storage area.

\section*{Acknowledgments}
We would like to thank Sonja Otten for proofreading the text and the
formulas and her valuable suggestions. We would like to thank Hans
Daduna for his help to create the formal mathematical model and Bernd
Heidergott for discussions of the model. We would like to thank the
Paderborn Center for Parallel Computing for providing their clusters
for our numerical experiments, Robert Lion Gottwald for suggestions
for MIP solvers and Nils Thieme for technical support.

\bibliography{psa}

\end{document}